\documentclass{article}

\usepackage{microtype}
\usepackage{graphicx}
\usepackage{subfigure}
\usepackage{booktabs} % for professional tables

\usepackage{natbib}
\usepackage[utf8]{inputenc} % allow utf-8 input
\usepackage[T1]{fontenc}    % use 8-bit T1 fonts
\usepackage{url}    % simple URL typesetting
\usepackage{amsfonts}       % blackboard math symbols
\usepackage{nicefrac}       % compact symbols for 1/2, etc.
\usepackage{microtype}      % microtypography
\usepackage{mathtools}
\usepackage[percent]{overpic}  % superimpose text over figures

\newtheorem{theorem}{Theorem}
\newtheorem{lemma}{Lemma}

\newtheorem{proof}{Proof}

\usepackage{algcompatible}
\usepackage{algpseudocode}
\usepackage{algorithm}

\newcommand{\comment}[1]{}
\usepackage{multirow}
\usepackage[usenames,dvipsnames]{xcolor}

\usepackage[colorlinks,linktoc=all]{hyperref}
 \usepackage[all]{hypcap}
 \hypersetup{citecolor=MidnightBlue}
 \hypersetup{linkcolor=MidnightBlue}
 \hypersetup{urlcolor=MidnightBlue}

\usepackage[nameinlink]{cleveref}

\graphicspath{ {./figures/} }
\usepackage[accepted]{icml2021}

\icmltitlerunning{Discover}

\begin{document}

\twocolumn[
\icmltitle{Variance Reduction in Deep Learning: More Momentum is All You Need.}




\begin{icmlauthorlist}
\icmlauthor{Lionel Tondji}{tubs,aires}
\icmlauthor{Sergii Kashubin}{goo}
\icmlauthor{Moustapha Cisse}{goo}
\end{icmlauthorlist}

\icmlaffiliation{aires}{This work was done during the AI Resideny at Google Brain}
\icmlaffiliation{goo}{Google Research, Brain Team}
\icmlaffiliation{tubs}{Institute of Analysis and Algebra, TU Braunschweig}
\icmlcorrespondingauthor{Lionel Tondji}{l.ngoupeyou-tondji@tu-braunschweig.de}

\vskip 0.3in
]



\printAffiliationsAndNotice{} % otherwise use the standard text.

\begin{abstract}
Variance reduction (VR) techniques have contributed significantly to accelerating learning with massive datasets in the smooth and strongly convex setting~\cite{schmidt2017minimizing, johnson2013accelerating, roux2012stochastic}. However, such techniques have not yet met the same success in the realm of large-scale deep learning due to various factors such as the use of data augmentation or regularization methods like dropout~\cite{defazio2019ineffectiveness}. 
   This challenge has recently motivated the design of novel variance reduction techniques tailored explicitly for deep learning~\cite{arnold2019reducing, ma2018quasi}. This work is an additional step in this direction. In particular, we exploit the ubiquitous clustering structure of rich datasets used in deep learning to design a family of scalable variance reduced optimization procedures by combining existing optimizers (e.g., SGD+Momentum, Quasi Hyperbolic Momentum, Implicit Gradient Transport) with a multi-momentum strategy~\cite{yuan2019cover}. Our proposal leads to faster convergence than vanilla methods on standard benchmark datasets (e.g., CIFAR and ImageNet). It is robust to label noise and amenable to distributed optimization. We provide a parallel implementation in JAX.
\end{abstract}

\section{Introduction}

Given a data generating distribution $\mathbb{P}$, a parameterized function $f_\theta$ (e.g. a deep network) and a loss function $\ell(\cdot)$, we consider the traditional risk minimization problem representative of most settings in ML~\cite{shalev2014understanding}:
\begin{equation}
\label{rm}
    \theta^* = \arg\min_{\theta}\mathbb{E}_{(x\sim \mathbb{P})}\ell(f_\theta(x))
\end{equation}

The celebrated optimization algorithm for solving this problem in large-scale machine learning is Stochastic Gradient Descent (SGD)~\cite{Robbins2007ASA,bottou2018optimization}. In favorable cases (e.g., smooth and strongly convex functions), SGD converges to a good solution by iteratively applying the following first-order update rule: $\theta_{t+1} \leftarrow \theta_t - \mu_t g(x_t, \theta_t)$ where $x_t$ is an instance drawn from $\mathbb{P}$, $\mu_t$ is the learning rate and $g(x_t, \theta_t)$ is the (approximate) gradient. SGD enjoys several desirable properties. Indeed, It is fast: when one has access to the actual gradient, its convergence rate is $O(e^{-\nu t})$ for some $\nu>0$. Also, it is memory efficient, robust, and amenable to distributed optimization~\cite{bottou2018optimization,nemirovski2009robust,dean2012large}. However, we generally do not know the data-generating distribution $\mathbb{P}$; consequently, we do not have access to the actual gradient. Instead, we resort to empirical risk minimization and rely on a stochastic approximation of the gradient for a given $x_t$ and $\theta_t$. Unfortunately, the gradient noise due to using the approximate instead of the exact gradient slows down the convergence speed (which becomes $O(1/t)$ instead of $O(e^{-\nu t})$)~\cite{bottou2018optimization}. It also makes the algorithm harder to tune. Variance Reduction (VR) techniques allow us to mitigate the impact of using noisy gradients in stochastic optimization~\cite{roux2012stochastic,defazio2014saga,johnson2013accelerating,mairal2013optimization}.

Variance reduction methods have been successfully applied to convex learning problems~\cite{roux2012stochastic,defazio2014saga,shalev2013stochastic,johnson2013accelerating}. However, this success does not readily translate to large-scale non-convex settings such as deep learning due to colossal memory requirements~\cite{roux2012stochastic,defazio2014saga,shalev2013stochastic} and the use of normalization and data augmentation procedures~\cite{defazio2019ineffectiveness}. This has motivated the design of novel VR strategies that are better suited for deep learning~\cite{cutkosky2019momentum,arnold2019reducing,ma2018quasi}. In the next section, we present in more detail \emph{Implicit Gradient Transport}~\cite{arnold2019reducing} and \emph{Quasi Hyperbolic Momentum}~\cite{ma2018quasi}, two instances of successful application of VR to deep learning.

Large scale datasets used in deep learning problems come with a rich clustering structure~\cite{deng2009imagenet}. For example, the data can result from the aggregation of several datasets collected by different individuals or organizations (e.g., this is typical in the federated learning setting~\cite{li2019federated}). When there is an underlying clustering structure in the data, the variance due to gradient noise decomposes into an in-cluster variance and a  between-cluster variance. ~\citet{yuan2019cover} exploit this insight to reduce the between cluster variance in the convex setting by considering every data point as a separate cluster. Unfortunately, such an approach requires storing as many models as data points and therefore has prohibitive memory requirements. Also, it does not consider the use of mini-batches during training. Consequently, it is not directly applicable to deep learning. 
\textbf{In this work, we make the following contributions}:
\begin{itemize}
\item We leverage the idea of \emph{using multiple momentum terms} and introduce a family of variance reduced optimizers for deep learning. These new approaches (termed Discover~\footnote{Deep SCalable Online Variance Reduction}) improve upon widely used methods such as SGD with Momentum, IGT, and QHM. They are theoretically well-motivated and can exploit the clustering structures ubiquitous in deep learning. 

\item Using simple clustering structures, we empirically validate that Discover optimizers lead to faster convergence and sometimes (significantly) improved generalization on standard benchmark datasets. We provide parallel implementations of our algorithms in JAX\footnote{https://github.com/google/jax}.
\end{itemize}

In the next section, we provide a brief overview of variance reduction methods with an emphasis on Implicit Gradient Transport~\cite{arnold2019reducing} and Quasi Hyperbolic Momentum~\cite{ma2018quasi}, two VR optimizers specifically designed for deep learning. We then discuss (section 3) the clustering structure present in most deep learning problems and how we can leverage it to design scalable variance reduction strategies. The experiments section demonstrates our proposals' effectiveness and provides several insights regarding their behavior.

\section{Variance Reduction in Deep Learning}

\emph{We consider Variance Reduction (VR) in the realm of large-scale deep learning~\cite{Goodfellow-et-al-2016}, i.e. when we are in the presence of significant amounts of (streaming) data, and the parameterized function $f_{\theta}$ is a massive deep neural network}. Most existing approaches do not naturally scale to this setting. Indeed, dual methods for VR, such as SDCA~\cite{shalev2013stochastic}, have prohibitive memory requirements due to the necessity of storing large dual iterates. Primal methods akin to SAGA~\cite{roux2012stochastic} are also memory inefficient because they need to store past gradients. Other approaches like SARAH~\cite{nguyen2017sarah} are computationally demanding; they entail a full snapshot gradient evaluation at each step and two minibatch evaluations. While some recent approaches, such as stochastic MISO~\cite{bietti2017stochastic}, can handle infinite data in theory. However, their memory requirement scales linearly with the number of examples. In addition to these limitations, Defazzio \& Bottou \cite{defazio2019ineffectiveness} have shown that several other factors such as data augmentation~\cite{krizhevsky2012imagenet}, batch normalization~\cite{ioffe2015batch}, or dropout~\cite{srivastava2014dropout} impede the success of variance reduction methods in deep learning. Despite these negative results, some recent approaches such as Quasi Hyperbolic Momentum~\cite{ma2018quasi} and Implicit Gradient Transport with tail averaging (IGT)~\cite{arnold2019reducing} have shown promising results in variance reduction in the context of large scale deep learning. In the sequel, we present them in more detail. We first present the Momentum or Heavy Ball method since the above approaches and our proposal rely on it.

\textbf{The Momentum or Heavy Ball}~\cite{polyak1964some,sutskever2013importance} method uses the following update rule:
\begin{align}
    v_t& =  \beta v_{t-1} + (1-\beta)\cdot g(\theta_t,x_t)  \nonumber \\
    \theta_{t+1}& =  \theta_t - \mu v_t \nonumber 
\end{align}
where $v_t$ is called the momentum buffer and $\beta_t$ balances the buffer and the approximate gradient compute at iteration $t$. SGD with Momentum is widely used in deep learning due to its simplicity and its robustness to variations of hyper-parameters~\cite{sutskever2013importance,zhang2017yellowfin}. It is an effective way of alleviating slow convergence due to curvature~\cite{goh2017momentum} and has been recently shown also perform variance reduction~\cite{roux2018online}.

\textbf{The Quasi Hyperbolic Momentum (QHM)}~\cite{ma2018quasi} method uses the following update rule:
\begin{align}
    v_t& =  \beta v_{t-1} + (1-\beta)\cdot g(\theta_t,x_t)  \nonumber \\
    \theta_{t+1}& =  \theta_t - \mu\cdot[\nu v_t + (1-\nu)g(\theta_t,x_t)] \nonumber
\end{align}
QHM extends Momentum by using in the update a combination of the approximate gradient and the momentum buffer. When $\nu=1$ (resp. $\nu=0$ ), it reduces to Momentum (resp. SGD). QHM inherits the efficiency of Momentum and also performs VR. In addition, it alleviates the potential staleness of the momentum buffer (when choosing $\nu < 1$). 

\textbf{The Implicit Gradient Transport (IGT)}~\cite{arnold2019reducing} method uses the following update rule:
\begin{align}
    \gamma_t& =  t/(t+1)  \nonumber \\
    v_t& =  \gamma_t \cdot v_{t-1} + (1-\gamma_t)\cdot g\bigg(\theta_t + \frac{\gamma_t}{1 - \gamma_t}(\theta_t - \theta_{t-1}),x_t \bigg)  \nonumber \\
    w_t& =  \beta \cdot w_{t-1} -\mu \cdot v_t  \nonumber \\
    \theta_{t+1}& =  \theta_t + w_t  \nonumber
\end{align}
IGT effectively achieves variance reduction by combining several strategies maintaining a buffer of past gradients (using iterative tail averaging) and transporting them to equivalent gradients at the current point. The resulting update rule can also be combined with Momentum as shown above.

\section{Exploiting Clusters in Variance Reduction}

\paragraph{The Ubiquitous Clustering Structure}
Large-scale machine learning datasets come with a rich clustering structure that arises in different ways depending on the setting. For example, when training deep neural networks, we combine various data augmentation strategies, resulting in a mixture with clusters defined by the transformations~\cite{zhang2017mixup,yun2019cutmix,krizhevsky2012imagenet}. In multiclass classification problems, one can consider each class as a separate cluster. Therefore, the overall dataset becomes a mixture defined by the classes. In all these cases, the data is not independent and identically distributed  across clusters. Indeed, different data augmentation strategies lead to different training data distributions. To make the clustering structure apparent, we can rewrite the minimization problem in equation~\ref{rm} as a combination of risks on the different clusters. To this end, we denote $\mathbb{P}_n$ the data distribution corresponding to the $n$-th cluster and  $p_n$ the probability with which we sample from that cluster, with $\sum_{i=1}^Np_n =1$. We also denote $x_t^{n}$ the realization of the data point $x_t$ belonging to the cluster $n$ for clarity. The risk minimization problem~\ref{rm} becomes:
\begin{equation}
\label{crm}
    \theta^* = \arg\min_{\theta}\sum_{n=1}^N p_n\mathbb{E}_{(x\sim \mathbb{P}_n)}\ell(f_\theta(x))
\end{equation} 
While one can pool all the data to apply traditional empirical risk minimization~\ref{rm}, this would ignore valuable \emph{prior clustering information}. In the next section, we show how, when solving the equivalent problem~\ref{crm}, we can leverage the clustering information to speed the learning procedure by reducing the variance due to gradient noise. We start with stochastic gradient descent and present a decomposition of such variance, which considers the clustering structure.  
\paragraph{Gradient noise for SGD with clusters} \label{sgd}
Here, we re-derive the variance of the gradient noise for SGD in presence of clustered data~\cite{sayed2014adaptive,yuan2019cover}. To this end, we introduce the filtration $F_t = \{\theta_{i<t+1}\}$ and denote $g_n(\theta) = \mathbb{E}_{(x^n\sim \mathbb{P}_n)}g(\theta,x^n)$ for convenience. We assume the loss $\ell(f(x^n))$ is $\delta$-Lipschitz with respect to $\theta$ and the clustered risk $\ell_n(\theta) = \mathbb{E}_{(x\sim \mathbb{P}_n)}\ell(f_\theta(x^n))$ is $\nu$-strongly convex, that is for any $\theta_1,\theta_2$ it holds: 
 $$(g_n(\theta_1) - g_n(\theta_2))^T(\theta_1-\theta_2)  \geq\nu\|\theta_1 -\theta_2 \|^2$$
For a given example $x_t^n$, the update rule for stochastic gradient descent is $\theta_{t+1} \leftarrow \theta_t - \mu_t g(x_t^n, \theta_t)$. The gradient noise resulting from this update depends both on the probability distribution on the clusters and the data distribution $\mathbb{P}_n$ for the considered cluster. For a given example, the gradient noise writes: $s_{t+1}(\theta_t) = g(x_t^n, \theta_t) - g(\theta_t)$ and the gradient noise \emph{within} the cluster $n$ is: $s_{t+1}^n(\theta_t) = g(x_t^n, \theta_t) - g_n(\theta_t)$. The following result bounds the first and second moment of the within-cluster variance of the gradient noise for SGD.

%$\mathbb{E}(s_{t+1}^n(\theta_t)|F_t = 0)$,
\begin{lemma}
\label{lm:bounds}
The first and second order moments of the gradient noise $s_{t+1}(\theta_t)$ satisfy: $\mathbb{E}\big(s_{t+1}(\theta_t)|F_t\big) = 0$ and $\mathbb{E}\big(\|s_{t+1}(\theta_t)\|^2|F_t\big) \leq \beta^2\|\Tilde{\theta}_t\|^2 + \sigma^2$ where $\beta^2 = 2 \delta^2$,  $\Tilde{\theta}_t = \theta_t - \theta^{*}$ and $\sigma^2 = 2 \mathbb{E}\big(\| g(x_t^n,\theta^{*})\|^2|F_t\big)$.
\end{lemma}
We can now decompose the variance of the gradient noise into a sum of  \emph{within-cluster} and \emph{between-cluster} variance. 

\begin{lemma}
\label{lm:sgd-variance-noise}
Assuming the gradient is unbiased and the variance of the gradient noise within the cluster $n$ is bounded as in shown in~\cref{lm:bounds}, the following inequality holds: 
\begin{equation}
    \mathbb{E}(\|s_{t+1}(\theta^*)\|^2|F_t) \leq \underbrace{\sum_{n=1}^Np_n\sigma^2_n}_{\tiny\mbox{in-cluster variance}} + \underbrace{\sum_{n=1}^Np_n\|g_n(\theta^*)\|^2}_{\tiny\mbox{between-cluster variance}}
    \label{eq:sgd-variance-noise}
\end{equation}
\end{lemma}
\Cref{lm:sgd-variance-noise} captures the structure of the problem. The LHS represents the variance of the gradient noise. The first term of the RHS is the within-cluster variance $\sigma_{in}^2$, and the second term is the between-cluster variance $\sigma_{bet}^2$. In the limit, the mean square deviation of the steady-state depends on the in-cluster and between-cluster variances as follows $\lim\sup_{t\rightarrow \infty} \|\theta_t - \theta^* \| = O(\mu(\sigma_{in}^2+\sigma_{bet}^2))$. Therefore when the clustering structure is known and $\sigma_{in}^2 \ll \sigma_{in}^2 + \sigma_{bet}^2$, which is our working assumption, we can significantly reduce the overall variance by reducing the between-cluster variance. In the next section, we present an update rule exploiting this fact and generalizing the approach presented in~\cite{yuan2019cover} to the minibatch setting\footnote{It is worth noting that \citet{yuan2019cover} have assumed the results in~\cref{lm:bounds} and state~\cref{lm:sgd-variance-noise} without proof. We provide full proofs in the appendix.}. We also prove its gradient noise properties and convergence\footnote{The proof assumes a smooth and strongly convex setting. Though we consider in the experiments non-convex problems, this assumption may be valid locally in a basin of attraction of the loss landscape of a deep neural network}

\section{Discover Algorithms}
To re-iterate, we assume we know the clustering structure and the probability $p_n$ of observing data from a given cluster $n$ such that $\sum_n p_n =1$. As we will show in the experiments, this is a realistic assumption, and straightforward design choices such as using labels or data augmentation strategies as clusters lead to improved results. We do not have access to the data distribution given a cluster $n$. We consider a learning setting where at each round $t$, we observe a batch of examples $\mathcal{B}_t=\{x_t^n\}$ coming from the different groups and sampled according to the mixture distribution induced by the clustering structure. We propose to achieve between cluster variance reduction in minibatch stochastic gradient descent by recursively applying the following  update rule: 
\begin{equation}
    \label{eq:discover-update}
    \theta_{t+1} = \theta_{t} - \frac{\mu}{|\mathcal{B}_t|} \cdot \sum_{x_t^n \in \mathcal{B}_t} \bigg(g(x_t^{n}, \theta_{t}) - g^{(n)}_{t} + \sum_{k=1}^{N}p_kg^{(k)}_{t}\bigg)
\end{equation}
The update rule stems from the traditional use of control variates for variance reduction~\cite{Fish96} with the additional trick to use one control variate for each cluster given that the clustering structure is known. In \Cref{eq:discover-update}, each $g_t^{(n)}$ is an approximation of the actual cluster gradient $g_{n}(\theta_t)$ to which the example $x_t^n$ belongs, and $\bar{g}_{t} = \sum_{k=1}^{N}p_k g_t^{(k)}$ is the average cluster gradient. The gradient noise resulting from the update rule \ref{eq:discover-update} depends both on the probability distribution on the clusters and the data distribution $\mathbb{P}_n$ of each considered cluster. It can be written as follows: %$$u_{t+1}(\theta_t) = $$
\begin{eqnarray*}
u_{t+1}(\theta_t) = \frac{1}{|\mathcal{B}_t|} \cdot \sum_{x_t^n \in \mathcal{B}_t} \bigg(g(x_t^{n}, \theta_{t}) - g^{(n)}_{t} + \bar{g}_{t}\bigg) - g(\theta_t) 
\end{eqnarray*} 
In the sequel, we show how a recursive application of the above update rule ultimately leads to reduced between-cluster variance of the gradient noise $u_{t+1}(\theta_t)$. Before, we first state a lemma exposing the properties of the gradient noise $u_{t+1}(\theta_t)$. Similarly to \Cref{lm:sgd-variance-noise}, this results highlights how the variance of the gradient noise  decomposes into an in-cluster and a between-cluster variance. We provide a proof of the lemma in the appendix~\Cref{ss:proof-lemma-1}.
%A version of this lemma for updates with single examples has been stated in~\cite{yuan2019cover}. 
\begin{lemma} (gradient noise properties) \label{lm:1}
Under the same assumptions as in \Cref{sgd}, for a batch of size $|\mathcal{B}_t|$ the gradient is unbiased $E[u_{t+1}(\theta_t) |F_t] = 0$. Denoting \, $\Tilde{\theta}_{t} \coloneqq \theta^* - \theta_t$, $C_1 = 4\delta^2 $ 
%$C_2 = \sum_{n=1}^N p_n\sigma_n^2$ 
and $\sigma_n^2 = 2 \cdot \mathbb{E}(\|g(x_t^n, \theta^*)\|^2)$ the variance of the gradient noise is bounded as: 
\begin{equation}
\begin{split}
    \mathbb{E}\Bigg(\|u_{t+1}(\theta_t)\|^2|F_t\Bigg) \leq \frac{1}{|\mathcal{B}_t|} \cdot  C_1\|\Tilde{\theta_t}\|^2 + \underbrace{\frac{1}{|\mathcal{B}_t|} \cdot\sum_{n=1}^N p_n\sigma_n^2}_{\tiny\mbox{in-cluster variance}} \\
    + \underbrace{\frac{2}{|\mathcal{B}_t|} \cdot \sum_{n}^N p_n \|g^{(n)}_{t} - g_n(\theta^*)\|^2}_{\tiny\mbox{between-cluster variance}} 
    \label{eq:discover-variance-noise}
\end{split}
\end{equation}

\end{lemma}  
\paragraph{Remark}The second and the last term of the right-hand side of this bound are respectively the within-cluster and the between-cluster variance. If the approximate cluster gradient converges to the actual one $g^{(n)}_{t}\rightarrow g_{n}(\theta^*)$ as $t\rightarrow\infty$, the between-cluster variance vanishes. It is worth noting that when we use the same approximate gradient buffer $g^{M}$ for all the clusters, the update rule~\ref{eq:discover-update} reduces to that of SGD with Momentum~\cite{polyak1964some,goh2017momentum}. Consequently, the between cluster variance term in the above bound becomes $(2/|\mathcal{B}_t|)\cdot \|g^M_{t} - g_n(\theta^*)\|^2$ and may vanish as $t\rightarrow\infty$. Therefore, Momentum also can perform  between-cluster variance reduction and can be seen as a special case of the method proposed here, albeit operating at a coarser level (maintaining one general approximate cluster gradient buffer instead of one for each cluster). Momentum has been mainly considered as a method for fighting curvature~\cite{goh2017momentum}. Recent work has shown its variance reduction capabilities in specific cases~\cite{roux2018online}. We show in our experiments that Momentum indeed performs between-cluster variance reduction. We now show that recursive applications of the update rule~\ref{eq:discover-update} indeed leads to vanishing between-cluster variance. The full proof of \Cref{th:1} is provided in the appendix \Cref{ss:proof-theorem-1}.

\begin{algorithm}[t]
\small
\caption{Discover}
\label{algo:discover}
\begin{algorithmic}[]
    \State Initialization{$: \bar{g}_0 = 0, \alpha \in (0, p_{\min}), g^{(n)}_0 = 0, \alpha_n = \alpha / p_n$}
    %\Comment{$p_{\min} = \min \{p_1, \dotsc , p_N\}$}
    \For{$t=0, \dotsc ,T-1$}
        \State Get the cluster indexes $\mathcal{C}$  in the current batch $\mathcal{B}_t$ and 
        \State update $\theta_{t+1}$, $\{g^{(n)}_{t+1}\}^{N}_{n=1}$ and $\bar{g}_{t+1}$:
        \State {$\theta_{t+1} = \theta_{t} - \frac{\mu}{|\mathcal{B}_t|} \cdot \sum_{x_t^n \in \mathcal{B}_t} \bigg(g(x_t^{n}, \theta_{t}) - g^{(n)}_{t} + \bar{g}_{t}\bigg)$}
        \For{$n \in \mathcal{C}$} \Comment{in parallel}
            \State {$\mathcal{B}_t^n = \{x_t^k \in \mathcal{B}_t \ \vert \ k = n \}$}
            \State {$g^{(n)}_{t+1} =
            (1 - \alpha_n) g_{t}^{(n)} + \frac{\alpha_n}{|\mathcal{B}_t^n|} \sum\limits_{x_t^n\in\mathcal{B}_t^n} g(x_t^{n}, \theta_{t})$}
        \EndFor
        \State {$g^{(n)}_{t+1} = g^{(n)}_{t}$} for each $n \notin C$
        \State {$\bar{g}_{t+1} = \bar{g}_{t} - \frac{\alpha}{|\mathcal{B}_t|} \sum_{x_t^n\in\mathcal{B}_t} \left( g_{t}^{(n)} - g(x_t^{n}, \theta_{t}) \right)$}
    \EndFor
    \State \algorithmicreturn{} $  \theta_{T}$ 
%\EndProcedure
\end{algorithmic}
 \vspace{-3pt}
\end{algorithm}
\begin{theorem}
\label{th:1}
Under the same assumptions as in \Cref{sgd}, we denote $|\mathcal{B}_t|$ as the batch size and $\sigma_{in}^2 = \sum_{n=1}^N p_n\sigma_n^2$ as the in-cluster variance with  $\sigma_n^2 = 2 \cdot \mathbb{E}(\|g(x_t^n, \theta^*)\|^2)$, $p_{\min} = \min\{p_1, \dotsc, p_N \}$. For any step size satisfying
$\mu \leq\min \Bigg\{ \frac{\nu|\mathcal{B}_t|}{3 \delta^2(|\mathcal{B}_t| + 5)}, \frac{\alpha}{6\nu} \Bigg\}$ where $\alpha \in (0, p_{\min})$ and $\gamma = \frac{3\mu^2}{\alpha|\mathcal{B}_t|}$, 
%%%%%%%%%%%%%%%%%%%%%%%%%%%%%%%%%%%%%%%%%%%%%%
the iterate $\theta_{t}$ from Discover~\ref{algo:discover} converge in expectation to the solution $\theta^{*}$ with 
%a linear rate w.r.t 
the contraction factor $q = 1-\mu \nu < 1$ with $G_0 = \sum_{n=1}^N p_n \|g_n(\theta^*)\|^2$.
%%%%%%%%%%%%%%%%%%%%%%%%%%%%%%%%%%%%%%%%%%%%%%
it holds: 
\begin{equation}
    \mathbb{E}(\|\theta_{t} - \theta^{*} \|^2) \leq  q^{t} \Bigg( \mathbb{E}(\|\Tilde{\theta}_{0}\|^2) + \gamma G_{0}\Bigg) + \frac{4\mu}{\nu|\mathcal{B}_{t-1}|}  \sigma^2_{in}
\end{equation}

\begin{equation}
    \limsup_{t \rightarrow +\infty}\mathbb{E}(\|\theta_{t+1} - \theta^{*} \|^2) = O(  \mu\cdot\sigma^2_{in}/|\mathcal{B}_t|)
\end{equation}
\end{theorem}

\Cref{th:1} shows that when the step size is small, Discover~\ref{algo:discover} eliminates the between-cluster variance in the limit, hence reducing the overall variance of the gradient noise to the in-cluster variance. Therefore, when the latter is significantly smaller than the between-cluster variance, Discover can be an effective variance reduction strategy. We show in the experiments section that when the clustering structure to exploit is carefully chosen, Discover leads to faster convergence and sometimes results in improved generalization thanks to the cluster information. We now show how to leverage the idea of using multiple momentum terms based on a given clustering to improve Implicit Gradient Transport~\cite{arnold2019reducing} and Quasi Hyperbolic Momentum~\cite{ma2018quasi} (both relying on a single momentum in their vanilla version). The update rules for such extensions (respectively called Discover-IGT and Discover-QHM) follow. Our experiments will demonstrate the effectiveness of these methods. 
\paragraph{Discover-IGT (D-IGT) update rule}:
\begin{align}
\gamma_t& =  t/(t+1)  \nonumber \\
v_t& =  \gamma_t \cdot v_{t-1} + (1-\gamma_t)\cdot g\bigg(\theta_t + \frac{\gamma_t}{1 - \gamma_t}(\theta_t - \theta_{t-1}),x_t \bigg)  \nonumber \\
    g^{(n)}_{t+1}& = \left\{
    \begin{array}{ll}
        (1 - \alpha_n) g_{t}^{(n)} + \alpha_n v_t  & \mbox{if } n \in \mathcal{C},\\
        g_{t}^{(n)} & \mbox{otherwise}
    \end{array} \right. \nonumber \\
        \theta_{t+1}& = \theta_{t} - \frac{\mu}{|\mathcal{B}_t|} \cdot \sum_{x_t^n \in \mathcal{B}_t} \bigg(v_t - g^{(n)}_{t} + \bar{g}_{t}\bigg) \nonumber
\end{align}

\textbf{The Discover-QHM (D-QHM)} method uses the following update rule:
\begin{align}
    g^{(n)}_{t+1}& = \left\{
    \begin{array}{ll}
        (1 - \alpha_n) g_{t}^{(n)} + \frac{\alpha_n}{|\mathcal{B}_t^n|} \sum\limits_{x_t^n\in\mathcal{B}_t^n}
        g(x_t^{n}, \theta_{t})  & \mbox{if } n \in \mathcal{C},\\
        g_{t}^{(n)} & \mbox{otherwise}
    \end{array} \right. \nonumber \\
    g^{(n)}_{t+2}& = \nu g^{(n)}_{t+1} + (1-\nu) \cdot \frac{1}{|\mathcal{B}_t^n|} \sum\limits_{x_t^n\in\mathcal{B}_t^n}
        g(x_t^{n}, \theta_{t}) \nonumber \\
        \theta_{t+1}& = \theta_{t} - \frac{\mu}{|\mathcal{B}_t|} \cdot \sum_{x_t^n \in \mathcal{B}_t} \bigg(g(x_t^{n}, \theta_{t}) - g^{(n)}_{t+1} + \bar{g}_{t+1}\bigg) \nonumber
\end{align}

where $\mathcal{B}_t^n = \{x_t^k \in \mathcal{B}_t \ \vert \ k = n \}$ is the batch at iteration $t$, $\mathcal{C}$ is the cluster indexes present in the batch $\mathcal{B}_t$, and $g^{(n)}_{t}$ are updated in parallel for $n \in \mathcal{C}$. 
Discover-IGT combines gradient transport and the multi-momentum strategy. D-QHM extends QHM and Discover by using in the update a combination of the approximate gradient and the momentum buffer, together with a multi-momentum approach similar to Discover. When $\nu=1$, and $N=1$ it reduces to Momentum. When $N=1$, it reduces to QHM. When $\nu=1$, it reduces to Discover.  Choosing $\nu < 1$ alleviates the potential staleness of the buffer.

\section{Experiments}

We validate that the Discover algorithms coupled with a careful clustering structure lead to faster convergence compared to their vanilla counterparts (SGD+Momentum, IGT, and QHM\footnote{we restrict QHM to $\nu \neq 0$ and $\nu \neq 1$ so it is not reduced to SGD or SGD+Momentum}). We provide additional insights and show that the runtime is comparable to competing methods thanks to a careful parallel implementation. 

We implement all the methods used in our experiments using JAX~\cite{jax2018github}, and FLAX~\cite{flax2020github} and perform our experiments on Google Cloud TPU~\cite{google_tpu} devices. We compare Discover with widely used minibatch SGD with Momentum, and Adam~\cite{kingma2014adam}, as well as with the more recently introduced IGT and QHM on ImageNet~\cite{deng2009imagenet} and CIFAR-10~\cite{krizhevsky2009learning} datasets. Discover is amenable to parallelization. For example, at each round, we can update the cluster gradients in parallel, as shown in \Cref{algo:discover}. Our implementation exploits this feature and groups clusters by cores to enable parallel updates. That is, examples of each core belong to the same cluster. In this section we treat $\alpha_n$ as an independent hyperparameter instead of using an exact equation $\alpha_n = \alpha / p_n$ to simplify the parallel implementation. We provide more details about the practical implementation in appendix \Cref{impl_details}. We confirm that using the same strategy does not make a practical difference for all the other optimizers we consider. For each optimizer, we select the hyperparameters either as suggested by its authors or by running a sweep to obtain the highest accuracy at the end of training across five random seeds.  The details of the training setup, HP search, and the best values found for each optimizer are given in appendix \Cref{impl_details_appendix_a}. 

\paragraph{ImageNet: Data augmentation methods  as clusters}
We first consider an image classification and train a ResNet-v1-50~\cite{He_2016_CVPR} model on the ImageNet \cite{deng2009imagenet, ILSVRC15} dataset for 
31200 steps (90 epochs). We use cosine learning rate schedule with mini-batches of a batch size of 4096, weight decay regularization of $0.001$, group normalization~\cite{Wu_2018_ECCV}, and weight standardization~\cite{qiao2019weight}. We use a standard preprocessing pipeline consisting of cropping~\cite{googlenet} with size 224x224, pixel value scaling, and a random horizontal flipping.

\begin{figure*}[ht!]
    \centering \footnotesize
    \begin{overpic}[width=0.31\linewidth,height=5.3cm,keepaspectratio]{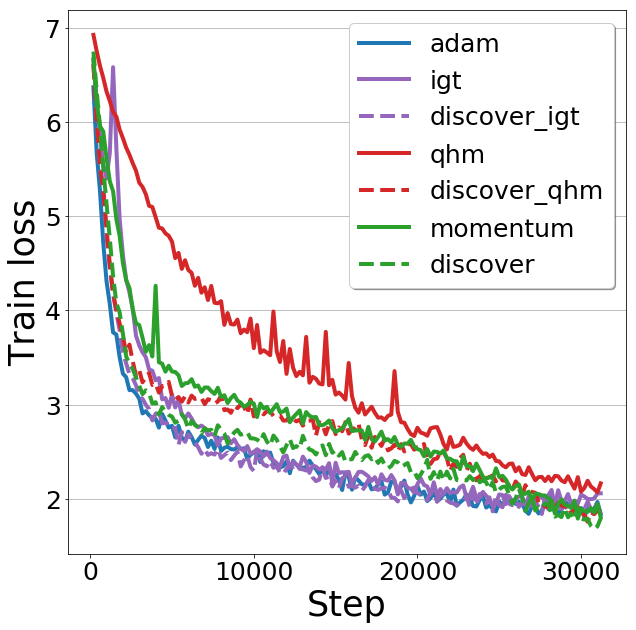}
        \put (20, 85) {(a)}
    \end{overpic}   
    \begin{overpic}[width=0.32\linewidth,height=5.3cm,keepaspectratio]{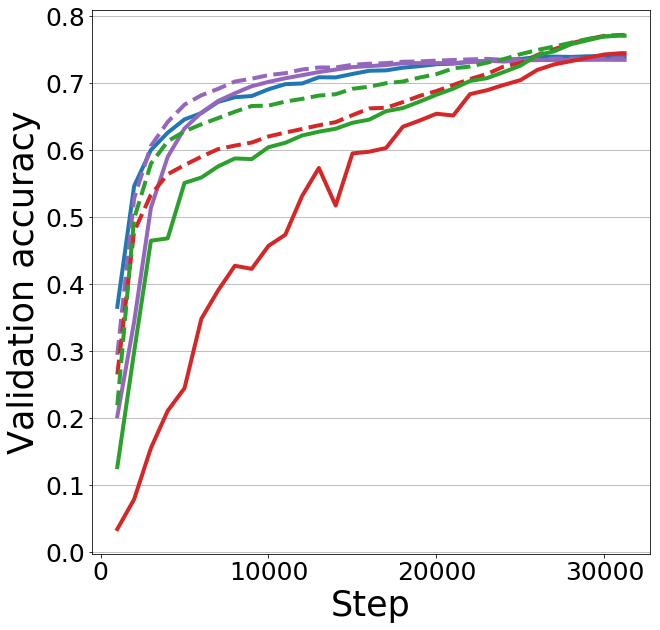}
        \put (20, 83) {(b)}
    \end{overpic}
    \begin{overpic}[width=0.33\linewidth,height=5.3cm,keepaspectratio]{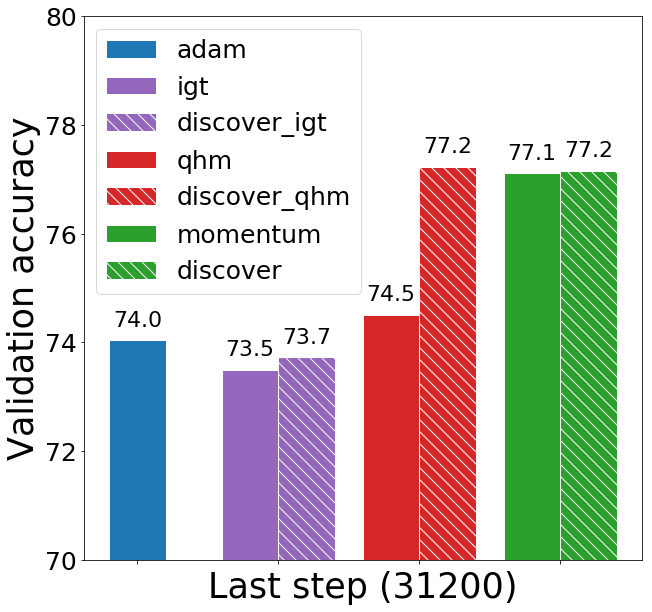}
        \put (82, 80) {(c)}
    \end{overpic}
    \smallskip
\caption{\footnotesize Results of training ResNet-v1-50 model on 
ImageNet dataset: train loss (a), validation accuracy (b) and validation accuracy on the last step in \% (c).
Discover consistently stays on par with vanilla optimizers (IGT, Momentum) or outperforms them (QHM) in the end of the training while always converging faster in the beginning.} \label{fig:mixup-plots}
\end{figure*}

\begin{figure*}[]
    \centering \footnotesize
    \begin{overpic}[width=0.28\linewidth,height=4.5cm,keepaspectratio]{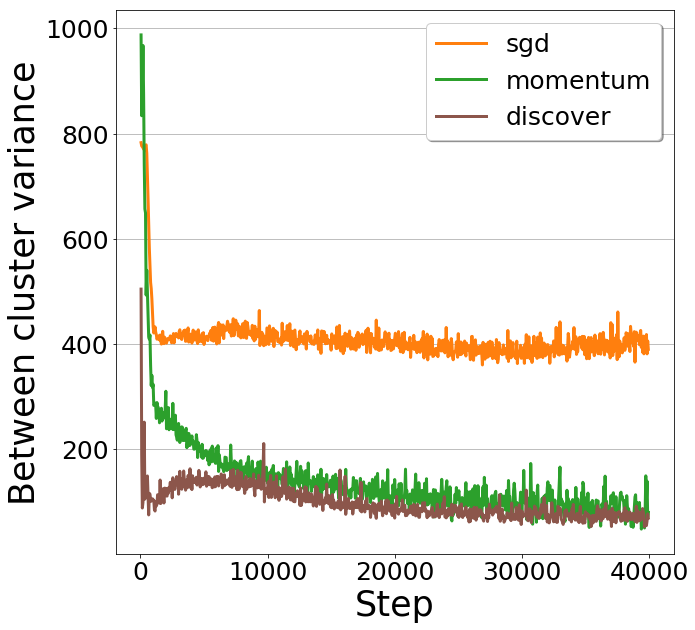}
        \put (25, 80) {(a)}
    \end{overpic}
    \begin{overpic}[width=0.28\linewidth,height=4.5cm,keepaspectratio]{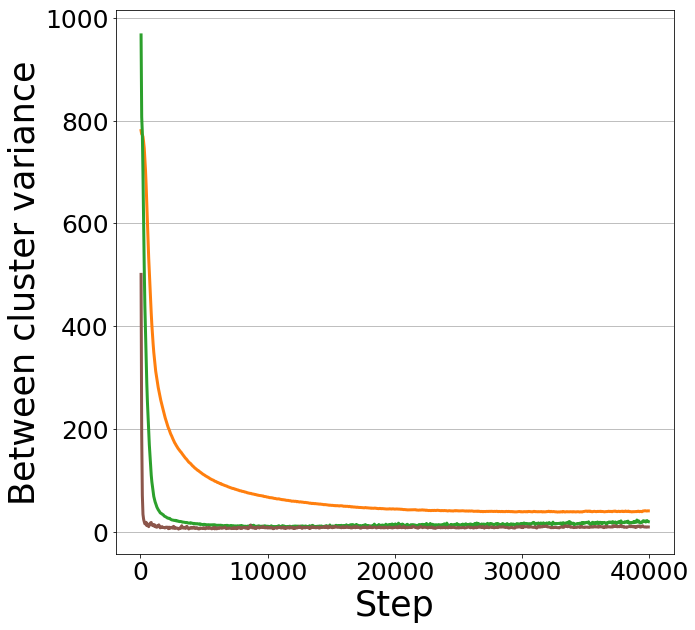}
        \put (25, 80) {(b)}
    \end{overpic}
    \begin{overpic}[width=0.42\linewidth,height=4.5cm,keepaspectratio]{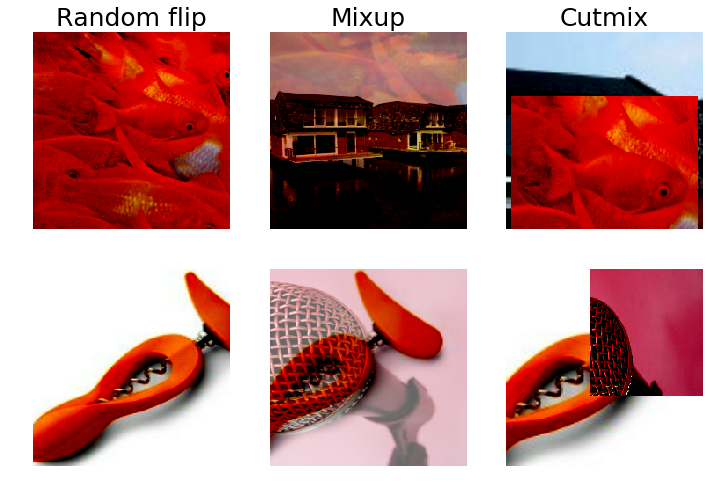}
        \put (5, 30) {(c)}
    \end{overpic}
\caption{\small Between-cluster variance estimate (mean per 100 train steps) for CIFAR-10 with clean labels (a) and noisy labels (b) with $p=0.8$. Example augmentations (c) used as clusters for ImageNet experiments.}
    \label{fig:gradient-noise-plots}
\end{figure*}

We train our Resnet-50 using three data augmentation methods: random flipping, Mixup~\cite{zhang2017mixup}, and CutMix~\cite{Yun_2019_ICCV}. For each image, we apply all transformations separately, producing three differently augmented examples (as shown in \Cref{fig:gradient-noise-plots}). \emph{Consequently, each of the augmentation methods induces a different cluster and the probability of each is $1/3$}. We compare the multi-momentum strategies (Discover, Discover-IGT, and Discover-QHM), exploiting this clustering information, with their vanilla counterparts (SGD with Momentum, IGT, and QHM). For the sake of completeness, we also add Adam to the comparison as a baseline. \Cref{fig:mixup-plots} shows the results for the different methods. Using all these data augmentation methods can make the learning problem more difficult because the resulting cluster data distributions can differ significantly. In this setting, we observe that Discover variants initially converge faster than their corresponding vanilla optimizers. Discover, and Discover-IGT reach similar final performance to SGD with Momentum and IGT, respectively, while Discover-QHM outperforms the vanilla QHM.

%\vspace{-10pt}

\begin{figure*}[]
    \centering \footnotesize
    \begin{overpic}[width=0.31\linewidth,height=5.3cm,keepaspectratio]{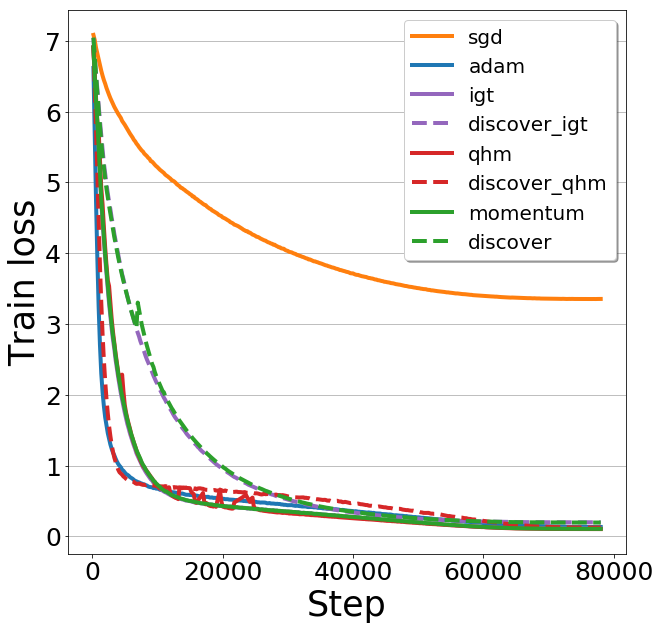}
        \put (22, 85) {(a)}
    \end{overpic}   
    \begin{overpic}[width=0.32\linewidth,height=5.3cm,keepaspectratio]{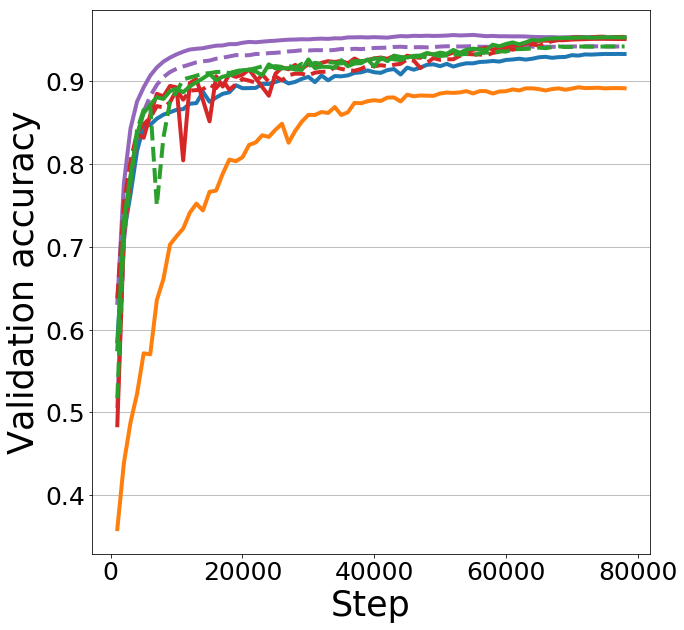}
        \put (82, 20) {(b)}
    \end{overpic}
    \begin{overpic}[width=0.33\linewidth,height=5.3cm,keepaspectratio]{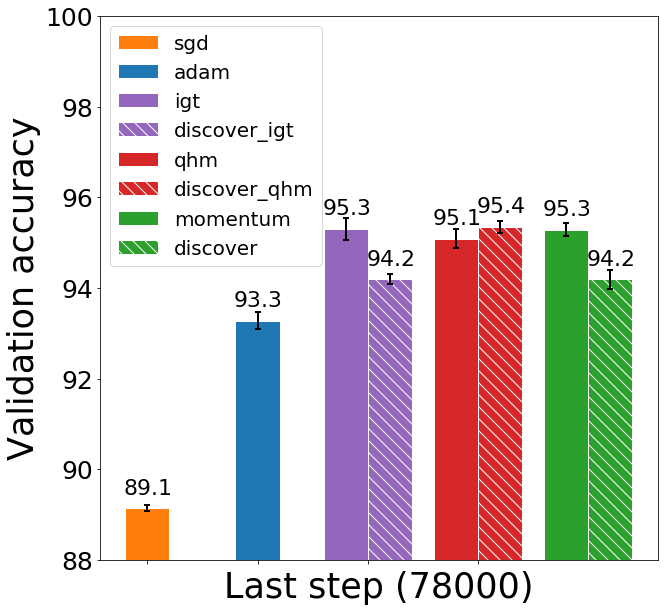}
        \put (82, 83) {(c)}
    \end{overpic}
    \smallskip
\caption{\footnotesize Results of training WideResNet26-10 model on CIFAR-10 dataset (classes as clusters): train loss (a), validation accuracy (b) and validation accuracy on the last step in \% (c). For each step a mean value across 5 random seeds is plotted, black whiskers in (c) indicate standard deviation. Discover variants for IGT and Momentum are slightly worse than vanilla optimizers, while being on par for QHM.}
    \label{fig:cifar-clean-data-plots}
\end{figure*}
%\vspace{-10pt}

\paragraph{CIFAR: Classes as clusters.}
Next, we consider the CIFAR-10~\cite{krizhevsky2009learning} classification and use the classes as the clusters, therefore having ten different clusters. We train a WideResNet26-10 \cite{zagoruyko2016wide} on 
CIFAR-10 for 400 epochs using cosine learning rate schedule, 
batch size of 256, group normalization, L2-regularization of $5\times10^{-4}$ and dropout rate of 0.3. The preprocessing consists of 4-pixel zero-padding, a random crop 
of size 32x32, 
scaling the pixels to $[0, 1]$ range and random horizontal flip.

\Cref{fig:cifar-clean-data-plots} shows the results for the different optimizers. CIFAR-10 is an easy task and all the methods eventually converge to high accuracy. IGT converges faster than Discover-IGT and reaches higher final accuracy ($95.3\%$ vs $94.2\%$). Discover (resp. Discover-QHM) converge similarly to SGD with Momentum (resp. QHM). These single momentum strategies also reach a similar (for QHM, $95.1\%$ vs $95.4\%$) or higher (for Momentum, $95.3\%$ vs $94.2\%$) final performance. To highlight the importance of carefully choosing the clustering structure, we also performed an experiment on CIFAR-10 where we assigned each point to 1 of 10 clusters \emph{uniformly at random} and train the model using discover and such clustering. The resulting test accuracy is $91.09\%$ down from $94.2\%$ when using classes as clusters. Therefore it is essential to use a good clustering structure.

%\vspace{-5pt}
\paragraph{Learning with noisy labels}
We now consider the more challenging ImageNet and CIFAR-10 classification task \emph{in the presence of label noise}: At each round, the label of the considered example is flipped with probability $p$ (the noise level) and assigned to a different class selected uniformly at random among the other classes. In this setting, the variance of the gradient noise is increased due to the presence of label noise (see Appendix \Cref{sn:variance-label-noise}).
We compare Discover, Discover-QHM, and Discover-IGT with their vanilla counterparts (SGD with Momentum, QHM, IGT) and Adam. We use Data augmentation methods as clusters for ImageNet and classes as clusters for CIFAR-10. We use the same setting as in the previous ImageNet and CIFAR-10 experiments. Still, We perform a hyperparameter (HP) search de novo to determine the best learning for each method in the presence of noisy labels. The details of the HP search, and the best values found for each optimizer are given in appendix \Cref{impl_details_appendix_a}.

\begin{figure*}[]
    \centering \footnotesize
    
    \begin{overpic}[width=0.31\linewidth,height=5.3cm,keepaspectratio]{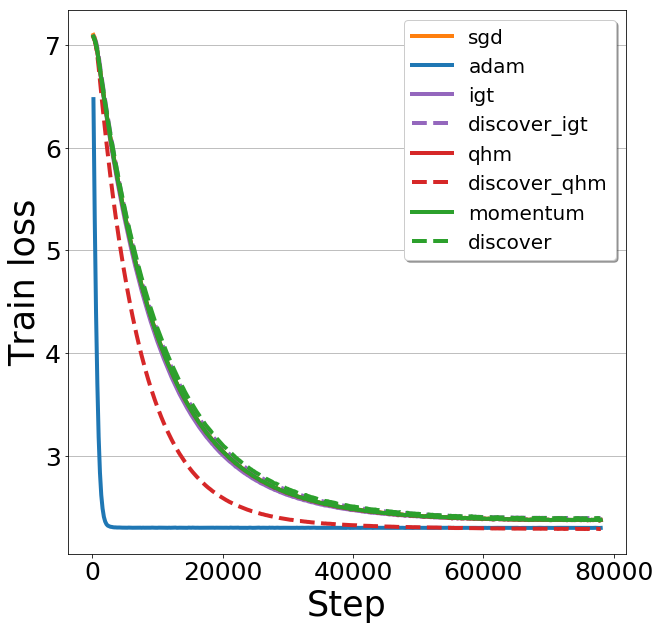}
        \put (20, 85) {(a)}
    \end{overpic}
    \begin{overpic}[width=0.32\linewidth,height=5.3cm,keepaspectratio]{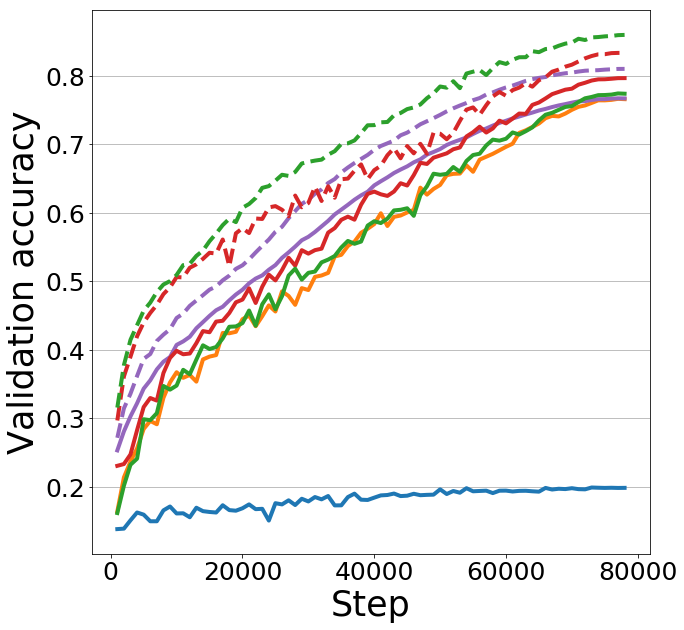}
        \put (20, 82) {(b)}
    \end{overpic}
    \begin{overpic}[width=0.32\linewidth,height=5.3cm,keepaspectratio]{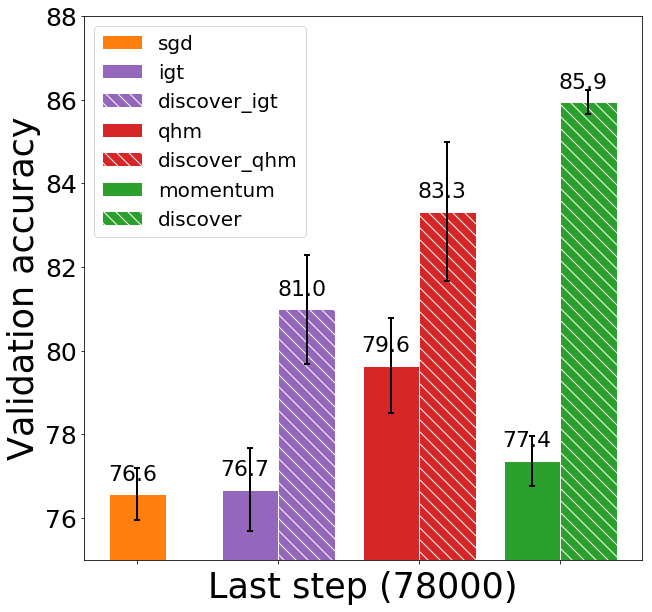}
        \put (70, 82) {(c)}
    \end{overpic}
    \smallskip
\caption{\small Results of training WideResNet26-10 model on CIFAR-10 dataset (classes as clusters) in a high noise setting $p=0.8$: train loss (a), validation accuracy (b) and validation accuracy on the last step in \% (c). For each step a mean value across 5 random seeds is plotted, black whiskers in (c) indicate standard deviation. Discover modifications outperform all vanilla optimizers by a large margin, once again suggesting high noise robustness. Adam achieves only 19.8\% mean (27.7\% max) final accuracy and thus not shown on (c) for clarity.}
    \label{fig:cifar-noisy-data-plots-08}
\end{figure*}

When the label noise is low $(p=0.2)$, all the methods achieve high accuracy on the non-corrupted training examples and the test examples with only a slight deterioration compared to a clean setting $p=0$ (exact results are given in appendix \Cref{cifar-noisy-appendix}). \Cref{fig:noisy-mixup-plots} and  \Cref{fig:cifar-noisy-data-plots-08}  show the results of this experiment in the high noise level setting $(p=0.8)$ on ImageNet and CIFAR-10, respectively. On CIFAR-10, the multi-momentum optimizers' loss curves are almost superimposed with the curves of SGD with Momentum, QHM, and IGT. However, the former generalize significantly better at each time step and reach higher final performance than all their single momentum counterparts. It is worth noting that Discover outperforms all the multi-momentum optimizers by a large margin on the validation accuracy, achieving $85.9\%$.
The superiority of multi-momentum optimizers over the single momentum ones translates to ImageNet, suggesting that our Discover algorithms find better local optima even in challenging settings. The other interesting finding is that while Adam converges faster in training loss, it generalizes poorly.

\begin{figure*}[]%[!ht]
    \centering \footnotesize
    \begin{overpic}[width=0.31\linewidth,height=5.3cm,keepaspectratio]{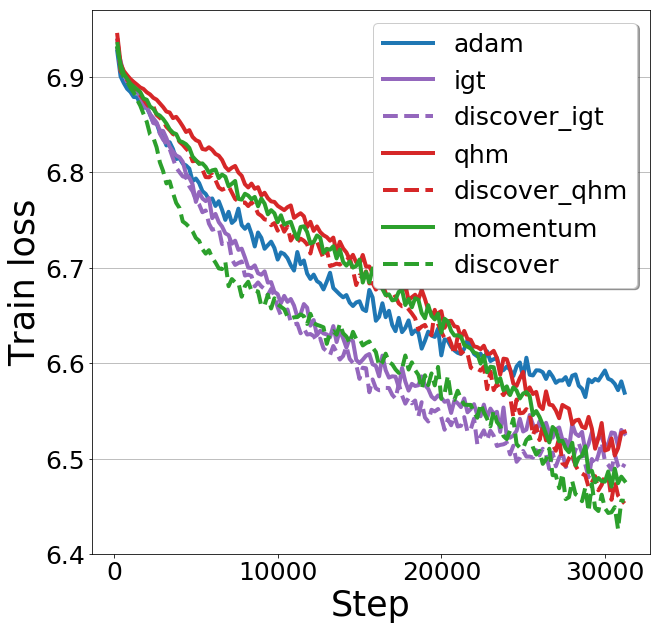}
        \put (20, 85) {(a)}
    \end{overpic}   
    \begin{overpic}[width=0.32\linewidth,height=5.3cm,keepaspectratio]{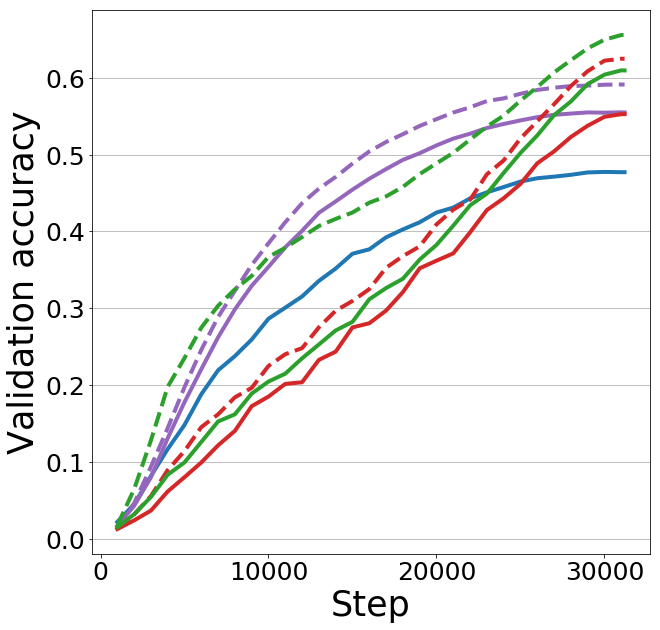}
        \put (20, 83) {(b)}
    \end{overpic}
    \begin{overpic}[width=0.33\linewidth,height=5.3cm,keepaspectratio]{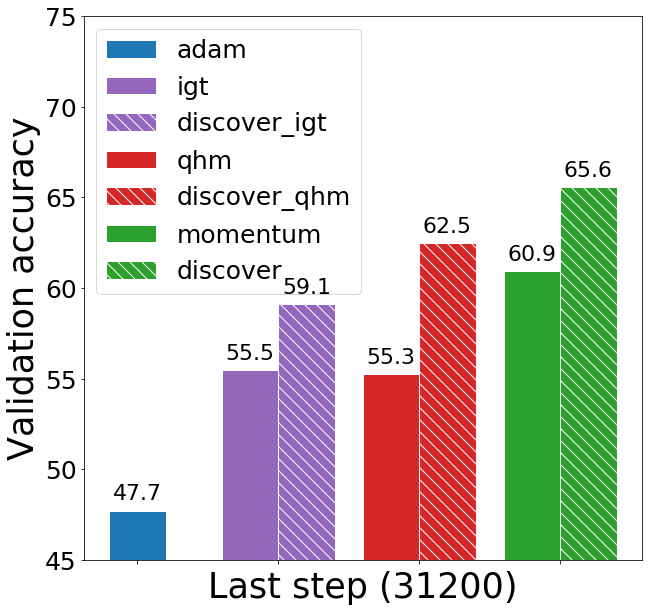}
        \put (82, 80) {(c)}
    \end{overpic}
    \smallskip
\caption{\small Results of training ResNet-v1-50 model on 
ImageNet dataset in a high noise setting $p=0.8$: train loss (a), validation accuracy (b) and validation accuracy on the last step in \% (c). Discover variants outperforms all vanilla optimizers by a large margin.} \label{fig:noisy-mixup-plots}
\end{figure*}
%\vspace{-10pt}

\begin{table}[t]
% \vspace{-2pt}
\footnotesize
  \caption{\footnotesize Training performance (in steps/second) with different optimizers. On ImageNet a single number is reported, while on CIFAR - Mean and std deviation across 5 runs.}
  \label{table:training-times}
  \centering
  \medskip
  \begin{tabular}[b]{lrr}
    \toprule
     \multirow{2}{*}{Optimizer} & \multicolumn{1}{c}{ImageNet,} & \multirow{1}{*}{CIFAR-10 ($\mu \pm \sigma), $} \\
    %\cmidrule(lr){2-3}
    \rule{0pt}{2ex} & 64 TPUv3 & 4 TPUv2 \\
    %{Optimizer} & {64 TPUv3, ImageNet} & {4 TPUv2, CIFAR-10 ($\mu \pm \sigma) $} \\
    \midrule
                          SGD       & $-.-$   & $14.3 \pm 3.8$    \\
                          Adam      & $8.2$   & $9.8 \pm 0.4$    \\
                          Momentum  & $8.7$   & $10.4 \pm 0.7$    \\
                          IGT       & $7.8$   & $9.3 \pm 0.6$ \\
                          QHM       & $8.6$   & $12.5 \pm 2.7$    \\
                          Discover  & $7.4$   & $6.3 \pm 0.3$    \\
                          Discover-QHM  & $7.6$   & $5.8 \pm 0.02$    \\
                          Discover-IGT  & $7.3$   & $5.6 \pm 0.02$    \\
    \bottomrule
  \end{tabular}
   %\vspace{-2pt}
\end{table}
%\medskip

\paragraph{Impact of clustering structure.} We also performed an experiment on CIFAR-10 where we compare three clustering structures: (1) Classes as clusters: This is the setting we have considered so far for CIFAR-10. (2) Transformations as clusters. (3) Random clusters: assigning each point to 1 of 10 clusters \emph{uniformly at random}. In each case, we train the model using Discover and the chosen clustering. When the clusters are selected uniformly at random, the resulting test accuracy is $91.09\%$. When we use transformation as clusters instead, the test accuracy improves to $93.6\%$. The best accuracy with Discover on CIFAR-10 ($94.2\%$) is achieved when using classes as clusters. However, using this clustering structure comes at the cost of a higher memory.

\paragraph{Between-Cluster Variance Reduction}
 %\vspace{-10pt}
In this experiment, we assess the between-cluster variance reduction capabilities of Discover compared to SGD and SGD with Momentum on CIFAR-10 (with classes as clusters) both in the clean $(p=0)$ and noisy $(p=0.8)$ settings. We measure at each step of the optimization an approximation of the between-cluster variance obtained by substituting $g_n(\theta_{t+1})$ to $g_n(\theta^*)$ in the between-cluster variance formula. For SGD we use \Cref{eq:sgd-variance-noise}. 
For Discover and SGD with Momentum --  \Cref{eq:discover-variance-noise}. However, for SGD with Momentum, one global gradient buffer is used for all the clusters.  \Cref{eq:discover-variance-noise} shows that in both the clean and the noisy setting, Discover reduces the between-cluster variance faster. The figure also highlights the between-cluster variance reduction capabilities of SGD with Momentum, which, though not as fast as Discover, also leads to quickly vanishing variance. This explains the good performance of Momentum in our previous experiments. 

% \vspace{-10pt}
\section{Conclusion and Perspectives}
 \vspace{-5pt}

We introduced a set of scalable VR algorithms exploiting the ubiquitous clustering structure in the data and relying on a multi-momentum strategy. We demonstrated that simple choices of clustering structure (e.g., transformations) can lead to improved results.
Our framework gives the designer significant flexibility allowing them to leverage prior information to select good clustering structure. The experiments demonstrated that the proposed strategy often leads to faster convergence and sometimes to improved generalization, especially in challenging settings (e.g., label noise).

 \Cref{table:training-times} shows there are only minor differences between the multi-momentum strategies and their vanilla counterparts thanks to our efficient implementation that exploits the parallel structure of the algorithms.  

\newpage
%\vspace{800mm}
%\newpage
%\pagebreak
%\pagebreak
%\newpage
\newpage

%\section*{Acknowledgements}

%\small
\bibliography{ms}
\bibliographystyle{icml2021}

\newpage 

\ % The empty page

\newpage

\section{Appendix A: Experimental details}
\label{impl_details_appendix_a}
\subsection{Implementation} \label{impl_details}
All models, training and evaluation pipelines are implemented using JAX ~\cite{jax2018github} and FLAX
~\cite{flax2020github}. The data loading and preprocessing is implemented in Tensorflow v2
~\cite{tensorflow2015-whitepaper} and TFDS ~\cite{TFDS}. The SGD, Momentum and Adam optimizers are taken from
FLAX libraries, QHM is implemented following \cite{ma2018quasi} and IGT is written in FLAX following an
example at \url{https://github.com/google-research/google-research/tree/master/igt\_optimizer}\
~\cite{arnold2019reducing}. Discover optimizers are implemented with single-program multiple-data training in mind. 
Our implementation expects a training data processed during a single training step by a given device core to
contain examples from the same cluster. After shuffling the dataset, our input pipeline selects the next
$batch\_size / num\_devices$ examples from the same randomly picked cluster to satisfy this condition. This
translates into a single cluster per batch on one-host one-core training (e.g. a workstation with one CPU),
but easily scales to more clusters by using multiple cores even within a single-host setting. For examples,
the CIFAR experiments use one host machine with 4 Google Cloud TPUv2~\cite{google_tpu} devices, thus having
8 cores in total, so each batch can contain examples from up to 8 different clusters. Discover optimizer
averages $g_{\Delta,t} = g(x_t^{n}, \theta_{t}) - g^{(n)}_{t}$ across devices to perform identical
$\bar{g}_{t}$ updates locally on each device and performs local $g^{(n)}_{t}$ updates in parallel, 
synchronizing across devices at the end of each training step (see \Cref{algo:discover} for
notation). This averaging makes effective $\alpha_n$ value depend on the distributed training setup and to account for it we chose to tune $\alpha_n$ as a direct hyperparameter instead of trying to incorporate the training setup into the formula from \Cref{algo:discover}.This implementation is well suited
to a good balance of clusters within a single global batch, which is the usual case for a small number of
clusters and a randomly shuffled dataset. We have not tested the performance when the cluster distribution
in a global batch is significantly skewed.

\subsection{Datasets}
The datasets used in this work are ImageNet-1000 ~\cite{deng2009imagenet, ILSVRC15} and CIFAR-10 
~\cite{krizhevsky2009learning}. The datasets were downloaded from 
\url{https://www.tensorflow.org/datasets/catalog/imagenet2012} and
\url{https://www.tensorflow.org/datasets/catalog/cifar10} respectively. The detailed information about 
the size and format of the train and validation splits are available at the corresponding web pages. 

\subsection{ImageNet (augmentations as clusters)} \label{imagenet-appendix}

We trained a ResNet-v1-50 ~\cite{He_2016_CVPR} model on the ImageNet ~\cite{deng2009imagenet, ILSVRC15} dataset for 31200
steps, corresponding to 100 epochs on the unaugmented dataset. We used:
\begin{itemize}
    \setlength\itemsep{-0.2em}
    \item Cosine learning rate schedule with 5 warmup steps.
    \item Batch size of 4096.
    \item Weight decay regularization setting of $\lambda=0.001$. We used
    decoupled weight decay inspired by~\cite{loshchilov2018decoupled} and each step update all network
    parameters by $\Theta_{t+1} = \Theta_{t} * (1 - \lambda \mu)$ where $\mu$ is the learning rate.
    \item Group normalization with 32 groups~\cite{Wu_2018_ECCV}.
    \item Weight standardization.
\end{itemize}
The preprocessing consists of Inception-style cropping
~\cite{googlenet} with size 224x224, scaling the pixel values to $[-1; 1]$ range and random horizontal flip.
In this setting the model achieves 0.76 accuracy with Momentum
~\cite{polyak1964some,sutskever2013importance} optimizer when $\mu = 0.1, \beta = 0.9$. We extend the
setting to use Mixup ~\cite{zhang2017mixup} and Cutmix ~\cite{Yun_2019_ICCV} augmentations instead of random
flipping, choosing the same mixing ratio for all examples in the single local batch of each distributed
training host. We correspondingly assign all examples augmented with random horizontal flip to cluster 0,
with Mixup - to cluster 1, Cutmix - to cluster 2. The clusters are used for Discover optimizer. 

The hyperparameters are selected from a hyperparameter sweep to obtain the highest mean accuracy at 31200 steps
across 5 random seeds. The learning rate $\mu$ is chosen from a set of \{0.03, 0.1, 0.3\} for all
optimizers. The further hyperparameters are selected from the following sets:
\begin{itemize}
    \setlength\itemsep{-0.2em}
    \item Discover: $\alpha, \alpha_n \in \{0.001, 0.01, 0.1, 0.9\}$
    \item Momentum: $\beta \in \{0.85, 0.9, 0.95, 0.99\}$
    \item QHM: $\beta = 0.999, \gamma = 0.7$ (as reported default for ImageNet in ~\cite{ma2018quasi})
    \item IGT: $\beta = 0.9, tail\_fraction = 90$ (as reported to be selected for ImageNet in ~\cite{arnold2019reducing})
    \item Adam: $\mu = 0.001, \beta_1 = 0.9, \beta_2 = 0.999 $
    \item Discover-QHM: $\alpha \in \{0.1, 0.6, 0.7, 0.8, 0.9, 0.99\}$, $\alpha_n \in \{0.01, 0.1, 0.9\}$, $\beta \in \{0, 1, 0.6, 0.7, 0.8\}$
    \item Discover-IGT: $\alpha \in \{0.1, 0.6, 0.7, 0.8, 0.9, 0.99\}$, $\alpha_n \in \{0.01, 0.1, 0.9\}$, tail\_fraction $\in \{18, 45, 50, 60, 90 , 180, 360\}$
\end{itemize}

The exact hyperparameters selected for each optimizer from the hyperparameters sweep are:
\begin{itemize}
    \setlength\itemsep{-0.2em}
    \item Discover: $\mu=0.1, \alpha=0.1, \alpha_n=0.01$
    \item Momentum: $\mu=0.1, \beta=0.9$
    \item QHM: $\mu=0.1$
    \item IGT: $\mu=0.1$
    \item Adam: $\mu = 0.001, \beta_1 = 0.9, \beta_2 = 0.999 $
    \item Discover-QHM: $\mu=0.1, \alpha = 0.9, \alpha_n = 0.1, \gamma = 0.9$
    \item Discover-IGT: $\mu=0.1, \alpha = 0.9, \alpha_n = 0.1,  \\ tail\_fraction = 180$
\end{itemize}

The training is setup in the multi-device data-parallel fashion on a pod of 8x8 Google Cloud TPUv3 ~\cite{google_tpu}. There are 16 host machines, each having access to 4 TPUv3 devices. The ImageNet
~\cite{deng2009imagenet, ILSVRC15} training and validation splits obtained from \url{https://www.tensorflow.org/datasets/catalog/imagenet2012} are divided into equal parts and each host has access to its own separate part. At most $num\_hosts = 16$ examples are excluded this way to ensure the equal division. The training examples inside each batch are arranged such that each device core receives only examples from a single cluster (see \Cref{impl_details}). We confirmed that this does not make
a practical difference for any other optimizer examined: the loss and accuracy curves look identical and the final accuracy is the same independent of whether this technique is applied or not.

\subsection{CIFAR (classes as clusters)} \label{cifar-clean-appendix}

We trained a WideResNet26-10 ~\cite{zagoruyko2016wide} model on the CIFAR10 dataset
~\cite{krizhevsky2009learning} for 400 epochs. We used:
\begin{itemize}
    \setlength\itemsep{-0.2em}
    \item Cosine learning rate schedule with 5 warmup steps.
    \item Batch size of 256.
    \item L2-regularization~\cite{l2_reg} set to 0.0005.
    \item Dropout~\cite{dropout} rate of 0.3. 
    \item Group normalization~\cite{Wu_2018_ECCV} with 16 groups in the first
layer in each block and 32 groups in the rest of the layers. 
\end{itemize}
The preprocessing consists of 4 pixel
zero-padding followed by a random crop of size 32x32, scaling the pixels to $[0; 1]$ range and random
horizontal flip. We correspondingly assign all examples with the same class to the same cluster (for
Discover optimizer). 

The hyperparameters are selected from a hyperparameter sweep to obtain the highest
mean accuracy at 400 epochs across 5 runs with different random seeds. The learning rate $\mu$ is chosen
from a set of \{0.001, 0.01, 0.03, 0.1, 0.175\} for all optimizers. The further hyperparameters are
selected from the following sets:
\begin{itemize}
    \setlength\itemsep{-0.2em}
    \item Discover: $\alpha, \alpha_n \in \{0.001, 0.01, 0.015, 0.02, \dots \\ 0.095, 0.1, 0.15, 0.2, 
                      \dots 0.95, 1.0\}$
    \item Momentum: $\beta = 0.9$ (as used for WideResNet on CIFAR-10 in \cite{zagoruyko2016wide}) 
    \item QHM: $\beta \in \{0.8, 0.9\}, \gamma \in \{0.1, 0.2, \dots, 0.8, 0.9\}$ 
    \item IGT: $\beta = 0.9, tail\_fraction = 18$ (as determined to be best for CIFAR-10 in 
                                                    ~\cite{arnold2019reducing}) 
    \item Adam: $\beta_1 = 0.9,  \beta_1 = 0.999, \epsilon = 1e-8$ (as a good default 
    settings reported  in \cite{kingma2014adam})
    \item Discover-QHM: $\alpha \in \{0.6, 0.7, 0.8, 0.9\}$, $\alpha_n \in \{0.01, 0.1, 0.9\}$, $\beta \in \{0.6, 0.7, 0.8, 0.9, 0.99\}$
    \item Discover-IGT: $\alpha \in \{0.095, 0.1, 0.6, 0.7, 0.8, 0.9, \\ 0.99\},\ \alpha_n \in \{0.01, 0.1, 0.9\}$, tail\_fraction $\in \{18, 45, \\ 50, 60, 90 , 180, 360\}$
\end{itemize}

The exact hyperparameters selected for each optimizer from the hyperparameters sweep are:
\begin{itemize}
    \setlength\itemsep{-0.2em}
    \item SGD: $\mu = 0.01$
    \item Discover: $\mu = 0.01, \alpha = 0.095, \alpha_n = 0.1$
    \item Momentum: $\mu = 0.03, \beta=0.9 $ 
    \item QHM: $\mu = 0.1, \gamma = 0.9, \beta = 0.9 $ 
    \item IGT: $\mu = 0.01, \beta=0.9, tail\_fraction=18 $ 
    \item Adam: $\mu = 0.001 $ 
    \item Discover-QHM: $\mu = 0.01, \alpha = 0.9, \alpha_n = 0.1, \beta=0.9$
    \item Discover-IGT: $\mu = 0.01, \alpha = 0.095, \alpha_n = 0.1, \\ tail\_fraction=18$
\end{itemize}

The training setup is the same as for ImageNet experiments (see appendix \Cref{imagenet-appendix}) but using only one host machine with 4 Google
cloud TPUv2 ~\cite{google_tpu}. The CIFAR-10 ~\cite{krizhevsky2009learning} training and test splits are
obtained from \url{https://www.tensorflow.org/datasets/catalog/cifar10}.

\subsection{CIFAR (nosiy labels)} \label{cifar-noisy-appendix}

\begin{figure*}[h]
    \centering \footnotesize
    
    \begin{overpic}[width=0.31\linewidth,height=5.3cm,keepaspectratio]{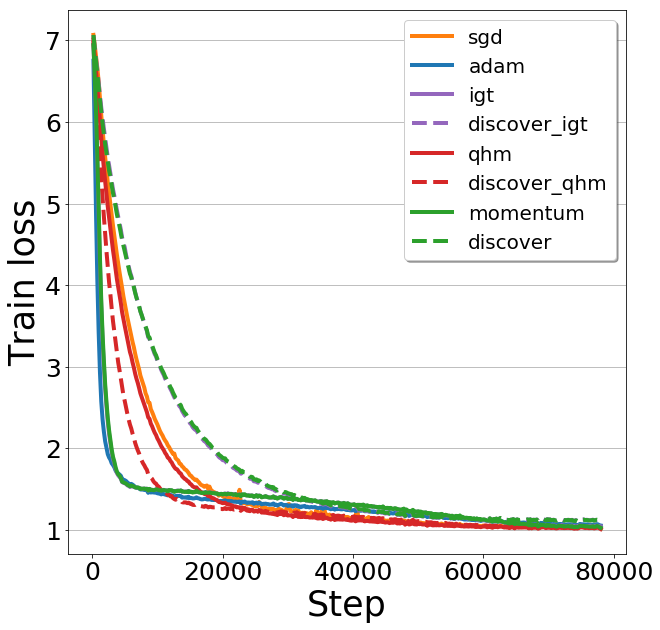}
        \put (20, 85) {(a)}
    \end{overpic}
    \begin{overpic}[width=0.32\linewidth,height=5.3cm,keepaspectratio]{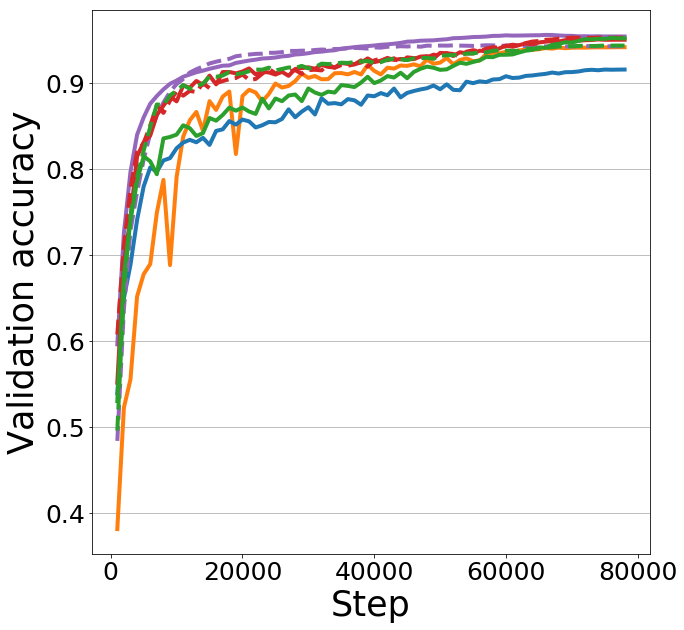}
        \put (20, 82) {(b)}
    \end{overpic}
    \begin{overpic}[width=0.32\linewidth,height=5.3cm,keepaspectratio]{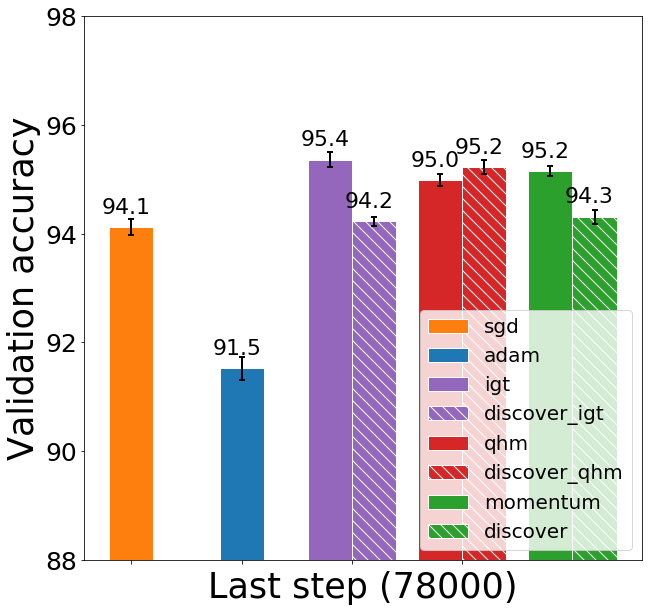}
        \put (70, 82) {(c)}
    \end{overpic}
    \smallskip
\caption{\small Results of training WideResNet26-10 model on CIFAR-10 dataset (classes as clusters) in a low noise setting $p=0.2$: train loss (a), validation accuracy (b) and validation accuracy on the last step in \% (c). For each step a mean value across 5 random seeds is plotted, black whiskers in (c) indicate standard deviation.}
    \label{fig:cifar-noisy-data-plots-02}
\end{figure*}

We also run the above-mentioned CIFAR-10 ~\cite{krizhevsky2009learning} experiments with partially corrupted
labels, where the label of each image is flipped to a random different class independently with probability
$p$ every time the example is seen during training. E.g. the same example might get different labels in different training epochs. 
We have tuned the hyperparameters anew in a same way as the clean CIFAR-10 experiment above \Cref{cifar-clean-appendix} each
value of $p \in \{0.2, 0.8\}$. The results for the low-noise setting ($p=0.2$) and high-noise setting ($p=0.8$) are reported in  \Cref{fig:cifar-noisy-data-plots-02} and \Cref{fig:cifar-noisy-data-plots-08} respectively.

The exact hyperparameters selected for $p=0.2$ setting are:
\begin{itemize}
    \setlength\itemsep{-0.2em}
    \item SGD: $\mu = 0.175$
    \item Discover: $\mu = 0.01, \alpha = 0.095, \alpha_n = 0.1$
    \item Momentum: $\mu = 0.1, \beta = 0.9 $ 
    \item QHM: $\mu = 0.175, \gamma=0.9, \beta=0.9$ 
    \item IGT: $\mu = 0.1, \beta=0.9, tail\_fraction=18 $ 
    \item Adam: $\mu = 0.001 $ 
    \item Discover-QHM: $\mu=0.1, \alpha = 0.9, \alpha_n = 0.1, \gamma = 0.7$
    \item Discover-IGT: $\mu=0.1, \alpha = 0.095, \alpha_n = 0.1,  tail\_fraction = 18$
\end{itemize}

The exact hyperparameters selected for $p=0.8$ setting are:
\begin{itemize}
    \setlength\itemsep{-0.2em}
    \item SGD: $\mu = 0.1$
    \item Discover: $\mu = 0.01, \alpha = 0.095, \alpha_n = 0.1$
    \item Momentum: $\mu = 0.01, \beta=0.9 $ 
    \item QHM: $\mu = 0.1, \gamma = 0.6, \beta = 0.9 $ 
    \item IGT: $\mu = 0.01, \beta=0.9, tail\_fraction=18 $ 
    \item Adam: $\mu = 0.001 $ 
    \item Discover-QHM: $\mu=0.1, \alpha = 0.6, \alpha_n = 0.1, \gamma = 0.6$
    \item Discover-IGT: $\mu=0.1, \alpha = 0.095, \alpha_n = 0.1,  tail\_fraction = 18$
\end{itemize}

\subsection{ImageNet (nosiy labels)} \label{imagenet-noisy-appendix}

We also run the above-mentioned ImageNet~\cite{deng2009imagenet, ILSVRC15} experiments with partially corrupted
labels, where the label of each image is flipped to a random different class independently with probability
$p=0.8$ every time the example is seen during training. E.g. the same example might get different labels in different training epochs. We used the same hyperparameters as selected for the clean ImageNet experiment, except the learning rate which was selected again for each optimizer value of $p$ as described in \Cref{imagenet-appendix}. The best
values were exactly the values chosen in \Cref{imagenet-appendix}.

\subsection{CIFAR (augmentations as clusters)} \label{cifar-aug-appendix}

\begin{figure*}[h]
    \centering \footnotesize
    
    \begin{overpic}[width=0.31\linewidth,height=5.3cm,keepaspectratio]{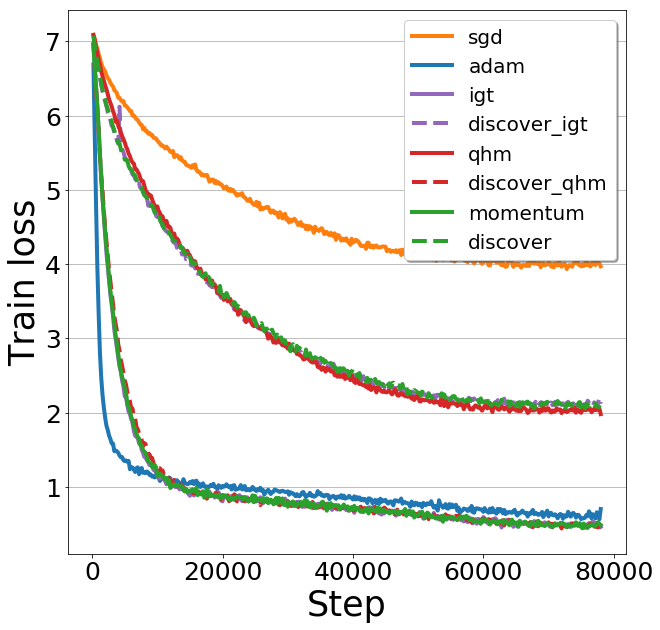}
        \put (20, 85) {(a)}
    \end{overpic}
    \begin{overpic}[width=0.32\linewidth,height=5.3cm,keepaspectratio]{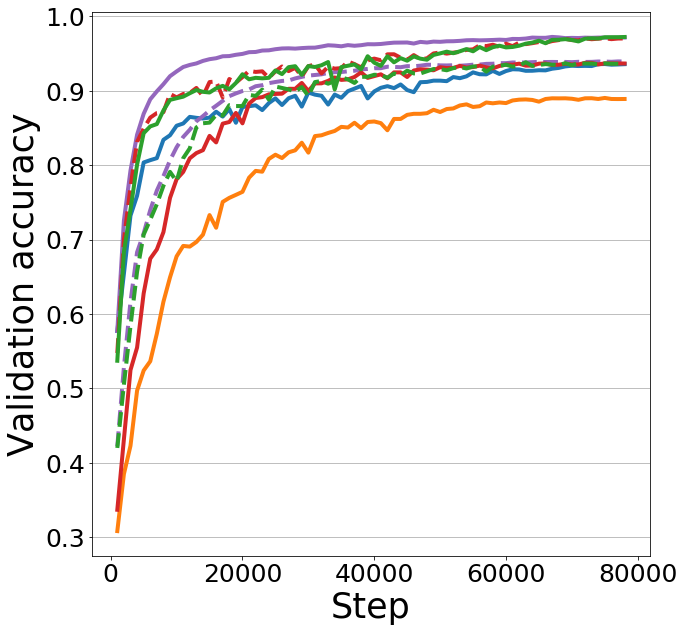}
        \put (20, 82) {(b)}
    \end{overpic}
    \begin{overpic}[width=0.32\linewidth,height=5.3cm,keepaspectratio]{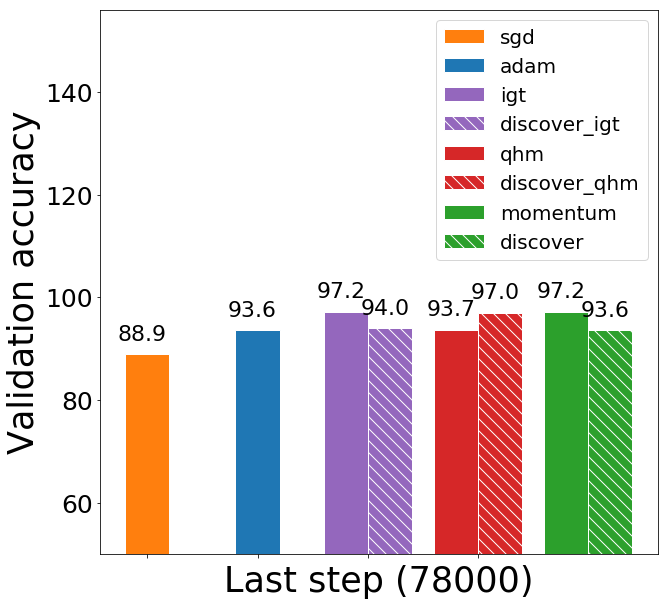}
        \put (20, 82) {(c)}
    \end{overpic}
    \smallskip
\caption{\small Results of training WideResNet26-10 model on CIFAR-10 dataset (Mixup, Cutmix and left-right flipping augmented examples as clusters) in a clean labels setting: train loss (a), validation accuracy (b) and validation accuracy on the last step in \% (c).} \label{fig:cifar-aug-clean-plots}
\end{figure*}

\begin{figure*}[h]
    \centering \footnotesize
    
    \begin{overpic}[width=0.31\linewidth,height=5.3cm,keepaspectratio]{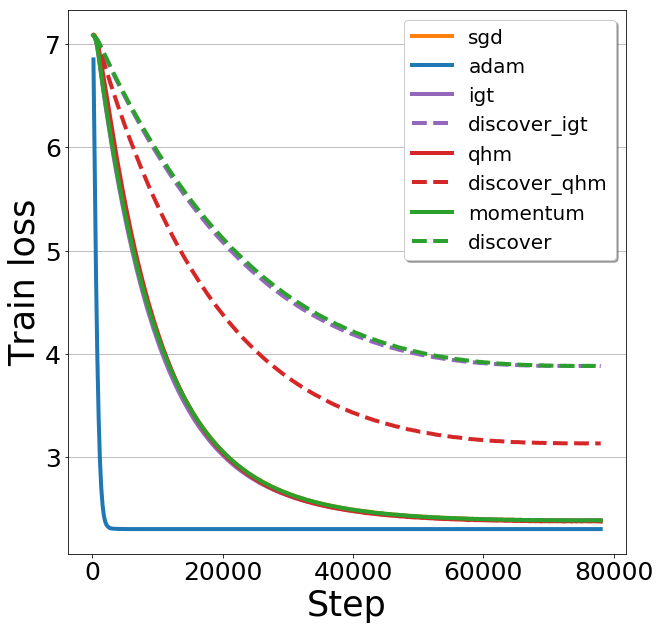}
        \put (20, 85) {(a)}
    \end{overpic}
    \begin{overpic}[width=0.32\linewidth,height=5.3cm,keepaspectratio]{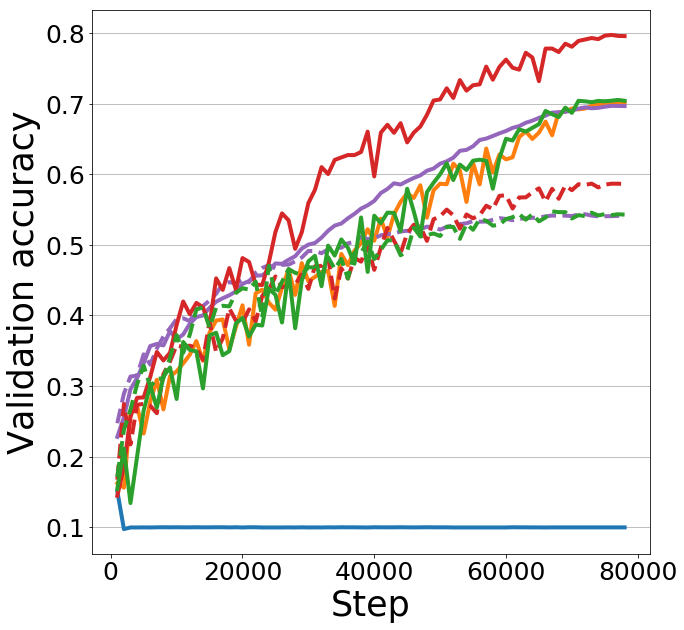}
        \put (20, 82) {(b)}
    \end{overpic}
    \begin{overpic}[width=0.32\linewidth,height=5.3cm,keepaspectratio]{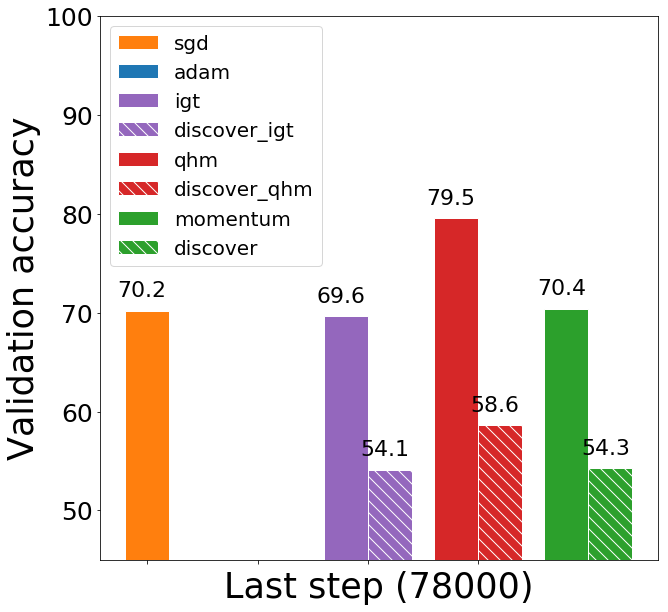}
        \put (70, 82) {(c)}
    \end{overpic}
    \smallskip
\caption{\small Results of training WideResNet26-10 model on CIFAR-10 dataset (Mixup, Cutmix and left-right flipping augmented examples as clusters) in a high noise ($p=0.8$) setting: train loss (a), validation accuracy (b) and validation accuracy on the last step in \% (c).} \label{fig:cifar-aug-08-plots}
\end{figure*}

We also run the above-mentioned CIFAR-10 experiments when using Mixup, Cutmix and left-right flipping augmented examples as clusters to mirror the setting used for ImageNet experiments above. We did not tune any hyperparameters anew, instead for both clean and noisy labels setting we have used the same hyperparameters as for the experiments above (when classes were used as clusters). When using clean labels the results for Discover modifications shown on \Cref{fig:cifar-aug-clean-plots} were very similar to those of vanilla optimizers as expected. When using noisy labels (with $p=0.8$) Discover modifications performed significantly worse than vanilla optimizers as shown on \Cref{fig:cifar-aug-08-plots}. These results highlight the importance of the careful clustering structure choice and suggest that Discover hyperparameters chosen for one clustering structure does not necessarily transfer to a different clustering choice in case of noisy labels.

\newpage

\section{Variance due to label noise}
\label{sn:variance-label-noise}

Suppose that we optimize the loss $\ell(f_\theta(x)) \coloneqq \ell(\theta; x, f_\theta(x))$ with gradient equal to $g(x, \theta)$. Let denote by $\Tilde{g}(x, \theta)$ the gradient of the loss $\ell(F_\theta(x))$ where $F(x)$ is the true labeling function. On an instance $x$, we can write it as: $\ell(f_\theta(x)) = a + b $ , with $a = \ell(F_\theta(x))$ and $b = \ell(f_\theta(x)) - \ell(F_\theta(x))$ .
\begin{eqnarray*}
\mathbb{E}\|g(x, \theta)\|^2  &=& \mathbb{E}\|\Tilde{g}(x, \theta) + g(x, \theta) - \Tilde{g}(x, \theta)\|^2 \\
&\leq& 2 \mathbb{E} \|\Tilde{g}(x, \theta)\|^2 + 2 \mathbb{E} \| g(x, \theta) - \Tilde{g}(x, \theta)\|^2
\end{eqnarray*}
The first term is the traditional variance of the gradient noise, and the second term is due to the label noise the larger the noise the larger it is.

\section{Appendix B: Proofs}

\begin{proof}

Before proving, we denote $\mathbb{E} [.] = \mathbb{E}_{n,t} [.|x_t^n]$ and $\mathbb{E}_{t} = \mathbb{E} [.|F_t]$ where the superscript $n$ indicate the cluster index $x$ arises from.\\

In this part we want to give a proof of \Cref{lm:bounds}.
We want to show that the first and second order moments of the gradient noise $s_{t+1}(\theta_t)$ satisfy : 
\begin{align}
  \mathbb{E}\Bigg(s_{t+1}(\theta_t)|F_t\Bigg) &= 0 \\
  \mathbb{E}\Bigg(\|s_{t+1}(\theta_t)\|^2|F_t\Bigg)& \leq \beta^2\|\Tilde{\theta}_t\|^2 + \sigma^2
\end{align}
where $\Tilde{\theta}_t = \theta_t - \theta^{*}$, $\beta^2 = 2\delta^2$ and :
\begin{eqnarray*}
\sigma^2 &=& 2 \mathbb{E}\Bigg(\| g(x_t^n, \theta^{*})\|^2\Bigg) \\
&=& 2 \sum_{n=1}^{N} p_n \mathbb{E}_{t} \Bigg(\| g(x_t^n, \theta^{*})\|^2 \Bigg) \\
\end{eqnarray*}
where $\sigma^2$ is referred to as the magnitude of gradient noise.\\
In case all the clusters have the same probability that is $p_1 = p_2,..., p_N = \frac{1}{N}$, we have that: 
\begin{eqnarray*}
\sigma^2 &=& 2 \mathbb{E}\Bigg(\| g(x_t^n, \theta^{*})\|^2\Bigg) \\
&=& 2 \sum_{n=1}^{N} p_n \mathbb{E}_{t} \Bigg(\| g(x_t^n, \theta^{*})\|^2 \Bigg) \\
&=& 2 \mathbb{E}_{t} \Bigg(\| g(x_t, \theta^{*})\|^2 \Bigg) \\
\end{eqnarray*}
which correspond to the magnitude of gradient noise without considering the clustering structure.\\
we recall that the gradient noise for SGD is given by : $s_{t+1}(\theta_t) = g(x_t^n, \theta_t) - g(\theta_t)$.
\begin{eqnarray*}
   \mathbb{E}\Bigg(s_{t+1}(\theta_t)|F_t\Bigg) &=&  \mathbb{E}_{n,t} \Bigg( g(x_t^n, \theta_t)|F_t\Bigg) - g(\theta_t) \\
&=& \sum_{n=1}^{N}p_n g_n (\theta_t) - g(\theta_t) \\ &=& 0
\end{eqnarray*}
Where we use the fact that : 
\begin{eqnarray*}
g_n (\theta_t) &=& \mathbb{E}_{t} \Bigg( g(x_t^n, \theta_t)|F_t\Bigg) \\
g(\theta_t) &=& \mathbb{E}_{n} (g_n (\theta_t)) 
\end{eqnarray*}

Using the following Jensen's inequality :
\begin{eqnarray*}
\mathbb{E}(\|a + b\|^2) &=& 4 \mathbb{E}(\|\frac{1}{2}a + \frac{1}{2}b\|^2) \\
&\leq& 2 \mathbb{E}(\|a\|^2) + 2 \mathbb{E}(\|b\|^2) 
\end{eqnarray*}

It holds that : 
\begin{eqnarray*}
\mathbb{E}\Bigg(\|s_{t+1}(\theta_t)\|^2|F_t\Bigg) &=& \mathbb{E}\Bigg(\|g(x_t^n, \theta_t) - g(\theta_t)\|^2|F_t\Bigg) \\
&\leq& 2 \mathbb{E}\Bigg(\|g(x_t^n, \theta_t) - g(\theta_t) - g(x_t^n, \theta^{*})\|^2|F_t\Bigg)  \\ + 2 \mathbb{E}\Bigg(\| g(x_t^n, \theta^{*})\|^2|F_t\Bigg)
\end{eqnarray*}
Using the fact that for any random variable $x$, 
\begin{equation*}
    \mathbb{E}\|x - E[x]\|^{2} = \mathbb{E}\|x\|^{2} - \|\mathbb{E}[x]\|^{2}  \leq \mathbb{E}\|x\|^{2} 
\end{equation*}
and  $g(\theta^{*}) = 0$, we have :
\begin{eqnarray*}
\mathbb{E}\Bigg(\|g(x_t^n, \theta_t) - g(\theta_t) - g(x_t^n, \theta^{*})\|^2|F_t\Bigg) 
&\leq& \mathbb{E}\Bigg(\|g(x_t^n, \theta_t) - g(x_t^n, \theta^{*})\|^2|F_t\Bigg) \\
&\leq& \delta^2 \|\Tilde{\theta}_t\|^2
\end{eqnarray*}

\end{proof}

We recall that the gradient noise \emph{within} the cluster $n$ is given by: 
\begin{equation*}
    s_{t+1}^n(\theta_t) = g(x_t^n, \theta_t) - g_n(\theta_t)
\end{equation*}

%$\mathbb{E}(s_{t+1}^n(\theta_t)|F_t = 0)$,
Assuming the gradient noise \emph{within} the cluster $n$ is following assumptions similar to the result in \Cref{lm:bounds} meaning that it is unbiased and it variance is bounded : $\mathbb{E}(\|s_{t+1}^n(\theta_t)\|^2|F_t) \leq \gamma_n^2\|\Tilde{\theta}_t\|^2 + \sigma_n^2$, where $\Tilde{\theta_t} = \theta^* - \theta_t$, and $\gamma_n, \sigma_n >0$, where $\sigma_n^2$ is referred to as the magnitude of gradient noise in cluster n. we want to prove \Cref{lm:sgd-variance-noise}  :
\begin{equation}
    \mathbb{E}(\|s_{t+1}(\theta^*)\|^2|F_t) \leq \underbrace{\sum_{n=1}^Np_n\sigma^2_n}_{\tiny\mbox{in-cluster variance}} + \underbrace{\sum_{n=1}^Np_n\|g_n(\theta^*)\|^2}_{\tiny\mbox{between-cluster variance}}
\end{equation}

%\pagebreak
\newpage

\begin{proof}
\begin{eqnarray*}
\mathbb{E}(\|s_{t+1}(\theta^*)\|^2|F_t) &=& \mathbb{E}(\|g(x_t^n, \theta^*)\|^2|F_t) \\
&=&\mathbb{E}(\|g(x_t^n, \theta_t) - g_n(\theta^*) + g_n(\theta^*)\|^2|F_t) \\
&=&\mathbb{E}(\|s_{t+1}^n(\theta^*) + g_n(\theta^*) \|^2|F_t) \\
&=& \mathbb{E}(\|s_{t+1}^n(\theta^*)\|^2|F_t) + \mathbb{E}(\|g_n(\theta^*) \|^2|F_t) \\
&\leq& \sum_{n=1}^Np_n(\gamma_n^2\|0\|^2 + \sigma_n^2) + \\ \sum_{n=1}^Np_n\|g_n(\theta^*)\|^2\\
\mathbb{E}(\|s_{t+1}(\theta^*)\|^2|F_t)&\leq& \sum_{n=1}^Np_n\sigma_n^2 + \sum_{n=1}^Np_n\|g_n(\theta^*)\|^2
\end{eqnarray*}
where the expectation is also taken over the different clusters with probabilities $p_n$.
\end{proof}

\subsection{Proof of \Cref{lm:1}}
\label{ss:proof-lemma-1}

Under the same assumptions as in \Cref{sgd}, for a batch of size $|\mathcal{B}_t|$ the gradient is unbiased $E[u_{t+1}(\theta_t) |F_t] = 0$. Denoting $\Tilde{\theta}_{t} \coloneqq \theta^* - \theta_t$, $C_1 = 4\delta^2 $, \, $C_2 = \sum_{n=1}^N p_n\sigma_n^2$ and $\sigma_n^2 = 2 \cdot \mathbb{E}(\|g(x_t^n, \theta^*)\|^2)$.\\

The gradient noise of DISCOVER is given by the following: 
\begin{eqnarray*}
u_{t+1}(\theta_t) &=&   \frac{1}{|\mathcal{B}_t|} \cdot \sum_{x_t^{n} \in \mathcal{B}_t} \bigg(g(x_t^{n}, \theta_{t}) - g^{(n)}_{t} + \bar{g}_{t}- g(\theta_t)\bigg) \\
&=& \frac{1}{|\mathcal{B}_t|} \sum_{x_t^{n} \in \mathcal{B}_t} u_{t+1}^n(\theta_t)
\end{eqnarray*} 

where 
\begin{eqnarray*}
u_{t+1}^n(\theta_t) =   g(x_t^{n}, \theta_{t}) - g^{(n)}_{t} + \bar{g}_{t}- g(\theta_t)
\end{eqnarray*} 

To this end, we introduce the filtration $F_t = \{\theta_{i< t+1}\}$.\\
Since $\theta_t \in F_t$ and  the cluster $n$ is selected with probability $p_n$, it holds that
\begin{eqnarray*}
    E[u^n_{t+1}(\theta_t) |F_t] &=& \mathbb{E}\Bigg(g(x_t^{n}, \theta_{t}) - g^{(n)}_{t} + \bar{g}_{t}- g(\theta_t) |F_t\Bigg) \\
    &=& \mathbb{E}\Bigg(g(x_t^{n}, \theta_{t}) - g(\theta_t) |F_t\Bigg) + \mathbb{E}\Bigg(\bar{g}_{t}- g^{(n)}_{t}|F_t\Bigg) \\
    &=& \mathbb{E}\Bigg(s_{t+1}(\theta_t)|F_t\Bigg) + \bar{g}_{t} - \sum_{n=1}^{N} p_n g^{(n)}_{t}\\
    &=& 0
\end{eqnarray*}
Because $s_{t+1}(\theta_t) = g(x_t^n, \theta_t) - g(\theta_t)$ and $\bar{g}_t = \sum_{n=1}^N p_n g_t^{(n)}$.\\

From ~\cite{yuan2019cover} it holds that : 
\begin{eqnarray*}
\mathbb{E}\Bigg(\|  u_{t+1}^n(\theta_t)\|^2| F_t\Bigg)
&\leq&  C_1 \|\Tilde{\theta_t}\|^2 + C_2 + \\ 2\sum_{n=1}^N p_n \|g^{(n)}_{t} - g_n(\theta^*)\|^2
\end{eqnarray*}

Now we can give properties of DISCOVER gradient noise by proving \Cref{lm:1}:
\newpage

\begin{eqnarray*}
    E[u_{t+1}(\theta_t) |F_t] &=& \mathbb{E}\Bigg(\frac{1}{|\mathcal{B}_t|} \sum_{x_t^{n} \in \mathcal{B}_t} u_{t+1}^n(\theta_t) |F_t\Bigg) \\
    &=& \frac{1}{|\mathcal{B}_t|} \sum_{x_t^{n} \in \mathcal{B}_t} \mathbb{E}\Bigg( u_{t+1}^n(\theta_t)|F_t \Bigg) \\
    &=& 0
\end{eqnarray*}

\begin{eqnarray*}
    \mathbb{E}\Bigg(\|u_{t+1}(\theta_t)\|^2\Bigg|F_t\Bigg) &=&   \mathbb{E}\Bigg(\|\frac{1}{|\mathcal{B}_t|} \sum_{x_t^{n} \in \mathcal{B}_t} u_{t+1}^n(\theta_t)\|^2\Bigg| F_t\Bigg) \\
    &=& \frac{1}{|\mathcal{B}_t|^2} \cdot \mathbb{E}\Bigg(\| \sum_{x_t^{n} \in \mathcal{B}_t} u_{t+1}^n(\theta_t)\|^2\Bigg| F_t\Bigg) \\
    &=& \frac{1}{|\mathcal{B}_t|^2} \cdot \mathbb{E}\Bigg(\sum_{x_t^{n} \in \mathcal{B}_t}\Bigg\|  u_{t+1}^n(\theta_t) \Bigg\|^2\Bigg| F_t\Bigg) + \\ \frac{1}{|\mathcal{B}_t|^2} \cdot \mathbb{E}\Bigg(\sum_{x_t^{n} \in \mathcal{B}_t}\sum_{x_t^{m} \neq x_t^{n}}\Bigg(  u_{t+1}^n(\theta_t)\Bigg)^T  \Bigg(  u_{t+1}^m(\theta_t)\Bigg)\Bigg| F_t\Bigg)\\
    &=& \frac{1}{|\mathcal{B}_t|^2} \cdot \mathbb{E}\Bigg(\sum_{x_t^{n} \in \mathcal{B}_t}\Bigg\|  u_{t+1}^n(\theta_t)\Bigg\|^2\Bigg| F_t\Bigg) \\ 
    &=& \frac{1}{|\mathcal{B}_t|^2} \cdot \sum_{x_t^{n} \in \mathcal{B}_t} \mathbb{E}\Bigg(\Bigg\|  u_{t+1}^n(\theta_t)\Bigg\|^2\Bigg| F_t\Bigg)  
\end{eqnarray*}

In the above proof we use the fact that the gradient noise are independent from one cluster to another meaning that for $n \neq m$: 
\begin{eqnarray*}
\mathbb{E}\bigg(u_{t+1}^n(\theta_t)u_{t+1}^m(\theta_t)| F_t\bigg) = \mathbb{E}\bigg(u_{t+1}^n(\theta_t)| F_t\bigg) \mathbb{E}\bigg(u_{t+1}^m(\theta_t)| F_t\bigg) = 0
\end{eqnarray*}

So we have : 
\begin{eqnarray*}
\mathbb{E}\Bigg(\|u_{t+1}(\theta_t)\|^2|F_t\Bigg) &=& \frac{1}{|\mathcal{B}_t|^2} \cdot \sum_{x_t^{n} \in \mathcal{B}_t} \mathbb{E}\Bigg(\|  u_{t+1}^n(\theta_t)\|^2| F_t\Bigg) \\
&\leq& \frac{1}{|\mathcal{B}_t|^2} \cdot \sum_{x_t^{n} \in \mathcal{B}_t} \Bigg( C_1 \|\Tilde{\theta_t}\|^2 + C_2 + 2\sum_{n=1}^N p_n \|g^{(n)}_{t} - g_n(\theta^*)\|^2 \Bigg)  \\
&\leq& \frac{1}{|\mathcal{B}_t|} \cdot  C_1 \|\Tilde{\theta_t}\|^2 + \frac{1}{|\mathcal{B}_t|} \cdot C_2 + \frac{2}{|\mathcal{B}_t|} \cdot \sum_{n=1}^N p_n \|g^{(n)}_{t} - g_n(\theta^*)\|^2 
\end{eqnarray*}

\newpage

\vspace*{100cm}

\begin{lemma}
\label{lm:2}
Under the same assumptions as in \Cref{sgd}, for a batch of size $|\mathcal{B}_t|$, defining $G_t$ to be :
\begin{eqnarray*}
    G_t = \sum_{n=1}^N p_n \mathbb{E}\Bigg(\|g^{(n)}_{t} - g_n(\theta^*)\|^2\Bigg)
\end{eqnarray*}

The two inequalities hold:
 \begin{equation}
 \begin{split}
    \mathbb{E}\Bigg(\|\Tilde{\theta}_{t+1}\|^2\Bigg) \leq \Bigg(1 - 2 \mu \nu + \mu^2 (\delta^2+ \frac{C_1}{|\mathcal{B}_t|})\Bigg)\mathbb{E}(\|\Tilde{\theta}_{t}\|^2)  \\ + \frac{\mu^2}{|\mathcal{B}_t|} C_2  + \frac{2\mu^2}{|\mathcal{B}_t|} \cdot G_t \nonumber
  \end{split}
\end{equation}

\begin{equation}
G_{t+1} \leq (1-\alpha) \cdot G_{t} + 3\alpha\delta^2 \cdot \mathbb{E}(\|\Tilde{\theta}_{t}\|^2) + \alpha C_2
\end{equation}
\end{lemma}

%\newpage
\subsection{Proof of \Cref{lm:2}}
Before proving \Cref{lm:2}, we are going to use the following lemma from ~\cite{sayed2014adaptation}:
\begin{lemma}(Mean-value theorem: Real arguments)
\label{lm:3}\\

Consider a real-valued
and twice-differentiable function $g(z) \in \mathbb{R}$, where $z \in \mathbb{R}^M$ is real-valued.
Then for any M-dimensional vectors $z_0$ and $\Delta z$,  the following increment equalities hold:

\begin{equation}
     g(z_0 + \Delta z) -  g(z_0) = \Bigg(\int_{0}^{1} \nabla_{z}g(z_0 + t \Delta z) dt\Bigg) \Delta z
\end{equation}

\begin{equation}
    \nabla_{z}g(z_0 + \Delta z) -  \nabla_{z}g(z_0) = (\Delta z)^{T} \Bigg(\int_{0}^{1} \nabla_{z}^{2}g(z_0 + r \Delta z) dr\Bigg)
\end{equation}
\end{lemma}

From \Cref{lm:3}, denoting $\Tilde{\theta}_{t} \coloneqq \theta^* - \theta_t$, $g(\theta) = \nabla \ell(f(\theta, x^n))$, we have: 
\begin{eqnarray*}
    g(\theta_t) &=& -  \Bigg(\int_{0}^{1} \nabla_{\theta}g(\theta^{*} -t \Tilde{\theta}_{t}) dt\Bigg)\Tilde{\theta}_{t}\\
    &\coloneqq& - H_{t}\Tilde{\theta}_{t}
\end{eqnarray*}
where we are introducing the symmetric time-variant matrix which is defined in terms of the Hessian of the cost function.
\begin{eqnarray*}
    H_{t} \coloneqq  \int_{0}^{1} \nabla_{\theta}g(\theta^{*} -t \Tilde{\theta}_{t}) dt
\end{eqnarray*}

Considering DISCOVER recursion in \Cref{algo:discover} under the same assumptions as in \Cref{sgd}, for a batch of size $|\mathcal{B}_t|$ and using the particular result of \Cref{lm:3} we have:  
\begin{eqnarray*}
    \theta_{t+1} &=& \theta_{t} - \frac{\mu}{|\mathcal{B}_t|} \cdot \sum_{x_t^{n} \in \mathcal{B}_t} \bigg(g(x_t^{n}, \theta_{t}) - g^{(n)}_{t} + \bar{g}_{t}\bigg)\\
    -\theta_{t+1} &=& -\theta_{t} + \frac{\mu}{|\mathcal{B}_t|} \cdot \sum_{x_t^{n} \in \mathcal{B}_t} \bigg(g(x_t^{n}, \theta_{t}) - g^{(n)}_{t} + \bar{g}_{t}\bigg)\\
    &=& -\theta_{t} + \frac{\mu}{|\mathcal{B}_t|} \cdot \sum_{x_t^{n} \in \mathcal{B}_t} \bigg(g(x_t^{n}, \theta_{t}) - g^{(n)}_{t} + \bar{g}_{t}- g(\theta_t) \\ + g(\theta_t)\bigg)\\
    &=& -\theta_{t} + \frac{\mu}{|\mathcal{B}_t|} \cdot \sum_{x_t^{n} \in \mathcal{B}_t} \bigg(g(x_t^{n}, \theta_{t}) - g^{(n)}_{t} + \bar{g}_{t}- g(\theta_t) \bigg) \\ + \mu \cdot g(\theta_t)\\
    &=& - \theta_{t} +  \mu \cdot g(\theta_t) + \mu \cdot u_{t+1}(\theta_t)\\
    \Tilde{\theta}_{t+1} &=& \Tilde{\theta}_{t} - \mu \cdot H_t \cdot \Tilde{\theta}_{t} + \mu \cdot u_{t+1}(\theta_t)\\
    \Tilde{\theta}_{t+1} &=& (I - \mu \cdot H_t)\Tilde{\theta}_{t} + \mu \cdot u_{t+1}(\theta_t) \\
\end{eqnarray*}

Since :
$$0 < \nu I_{d} < \nabla g(\theta) < \delta I_{d}$$ where $d$ is the dimension of $\theta$ and $\nabla g(\theta)$ is the Hessian Matrix, we can derive: $$(1- \mu \delta)I_{d} < I_{d} - \mu H_{t} <  (1 - \mu\nu) I_{d}$$ for all $t$.  Using the fact that $I_{d} - \mu H_{t}$ is a symmetric matrix, we have that its 2-induced norm is equal to its spectral radius so that:
\begin{eqnarray*}
\|I - \mu \cdot H_t\|^2 &=& \bigg(\rho(I - \mu \cdot H_t)\bigg)^2\\
&\leq& max \Bigg\{(1 - \mu \delta)^2, (1 - \mu \nu)^2 \Bigg\}\\
&=& max \Bigg\{1 - 2\mu \delta + \mu^2 \delta^2, 1 - 2\mu \nu + \mu^2 \nu^2 \Bigg\}\\
\|I - \mu \cdot H_t\|^2&\leq& 1 - 2\mu \delta + \mu^2 \delta^2
\end{eqnarray*}

\newpage

%\vspace*{100cm}
We got the last inequality because $\nu \leq \delta$

Using the previous result, we have : 
\begin{eqnarray*}
\|\Tilde{\theta}_{t+1}\|^2 &\leq& \|I - \mu \cdot H_t\|^2 \cdot \|\Tilde{\theta}_{t}\|^2+ \mu^2\cdot \|u_{t+1}(\theta_t)\|^2\\
\mathbb{E}\Bigg(\|\Tilde{\theta}_{t+1}\|^2|F_t\Bigg) &\leq& \|I - \mu \cdot H_t\|^2 \cdot \mathbb{E}(\|\Tilde{\theta}_{t}\|^2|F_t) + \mu^2 \mathbb{E}\Bigg(\|u_{t+1}(\theta_t)\|^2|F_t\Bigg)\\
& \leq & (1 - 2 \mu \nu + \mu^2 \delta^2)\mathbb{E}(\|\Tilde{\theta}_{t}\|^2|F_t) + \mu^2 \Bigg( \frac{1}{|\mathcal{B}_t|} C_1 \mathbb{E}(\|\Tilde{\theta}_{t}\|^2|F_t) + \frac{1}{|\mathcal{B}_t|} C_2 + \frac{2}{|\mathcal{B}_t|} \sum_{n=1}^N p_n \|g^{(n)}_{t} - g_n(\theta^*)\|^2 \Bigg) \\
&\leq& \Bigg(1 - 2 \mu \nu + \mu^2 (\delta^2+ \frac{C_1}{|\mathcal{B}_t|})\Bigg)\mathbb{E}(\|\Tilde{\theta}_{t}\|^2|F_t) + \frac{\mu^2}{|\mathcal{B}_t|} C_2  + \frac{2\mu^2}{|\mathcal{B}_t|} \cdot \sum_{n=1}^N p_n \|g^{(n)}_{t} - g_n(\theta^*)\|^2  
\end{eqnarray*}

Then taking the expectation on both side give us: 
\begin{eqnarray*}
\mathbb{E}\Bigg(\|\Tilde{\theta}_{t+1}\|^2\Bigg) &\leq& \Bigg(1 - 2 \mu \nu + \mu^2 (\delta^2+ \frac{C_1}{|\mathcal{B}_t|})\Bigg)\mathbb{E}(\|\Tilde{\theta}_{t}\|^2) + \frac{\mu^2}{|\mathcal{B}_t|} C_2  + \frac{2\mu^2}{|\mathcal{B}_t|} \cdot \sum_{n=1}^N p_n \mathbb{E}(\|g^{(n)}_{t} - g_n(\theta^*)\|^2)
\end{eqnarray*}

which conclude the proof. The second equation of \Cref{lm:2} is obtained from Lemma 2 of \cite{yuan2019cover}.

\newpage

\vspace*{100cm}
\subsection{Proof of \Cref{th:1}}
\label{ss:proof-theorem-1}
From \Cref{lm:2}, we have : 

\begin{eqnarray*}
\mathbb{E}(\|\Tilde{\theta}_{t+1}\|^2) + \gamma G_{t+1} &\leq& \Bigg(1 - 2 \mu \nu + \mu^2 (\delta^2+ \frac{C_1}{|\mathcal{B}_t|})\Bigg)\mathbb{E}(\|\Tilde{\theta}_{t}\|^2) + \frac{\mu^2}{|\mathcal{B}_t|} C_2  + \frac{2\mu^2}{|\mathcal{B}_t|} \cdot G_t + \\ \gamma \Bigg((1-\alpha) \cdot G_{t} + 3\alpha\delta^2 \cdot \mathbb{E}(\|\Tilde{\theta}_{t}\|^2) + \alpha C_2\Bigg) \\
&=& \Bigg[1 - 2 \mu \nu + \mu^2 (\delta^2+ \frac{C_1}{|\mathcal{B}_t|}) + 3\alpha \gamma\delta^2\Bigg]\mathbb{E}(\|\Tilde{\theta}_{t}\|^2) + (\frac{\mu^2}{|\mathcal{B}_t|} + \alpha\gamma) C_2 + \\
(\frac{2\mu^2}{|\mathcal{B}_t|} + \gamma(1-\alpha)) \cdot G_t\\
&\leq& \Bigg[1 - 2 \mu \nu + K \Bigg]\mathbb{E}(\|\Tilde{\theta}_{t}\|^2) + (\frac{\mu^2}{|\mathcal{B}_t|} + \alpha\gamma) C_2 +
(\frac{2\mu^2}{|\mathcal{B}_t|} + \gamma(1-\alpha)) \cdot G_t\\
&\leq& \Bigg[1 - 2 \mu \nu + K \Bigg]\Bigg(\mathbb{E}(\|\Tilde{\theta}_{t}\|^2) + 
\frac{\bigg(\frac{2\mu^2}{|\mathcal{B}_t|} + \gamma(1-\alpha)\bigg)}{1 - 2 \mu \nu + K} \cdot G_t \Bigg)+ \\(\frac{\mu^2}{|\mathcal{B}_t|} + \alpha\gamma) C_2 \\ 
&\leq& \Bigg[1 - 2 \mu \nu + K \Bigg]\Bigg(\mathbb{E}(\|\Tilde{\theta}_{t}\|^2) + 
\frac{\bigg(\frac{2\mu^2}{|\mathcal{B}_t|} + \gamma(1-\alpha)\bigg)}{1 - 2 \mu \nu} \cdot G_t \Bigg)+ \\(\frac{\mu^2}{|\mathcal{B}_t|} + \alpha\gamma) C_2 \\ 
&\leq& (1 - \mu \nu )\Bigg(\mathbb{E}(\|\Tilde{\theta}_{t}\|^2) + 
\frac{\bigg(\frac{2\mu^2}{|\mathcal{B}_t|} + \gamma(1-\alpha)\bigg)}{1 - 2 \mu \nu} \cdot G_t \Bigg)+ (\frac{\mu^2}{|\mathcal{B}_t|} + \alpha\gamma) C_2 \\
&\leq& (1 - \mu \nu )\Bigg(\mathbb{E}(\|\Tilde{\theta}_{t}\|^2) + 
\gamma \cdot G_t \Bigg)+ (\frac{\mu^2}{|\mathcal{B}_t|} + \alpha\gamma) C_2 \\
\mathbb{E}(\|\Tilde{\theta}_{t+1}\|^2) + \gamma G_{t+1}&\leq& (1 - \mu \nu )\Bigg(\mathbb{E}(\|\Tilde{\theta}_{t}\|^2) + 
\gamma \cdot G_t \Bigg)+ \frac{4\mu^2}{|\mathcal{B}_t|} \cdot C_2 
\end{eqnarray*}

\newpage

\vspace*{100cm}
where $K = 3\delta^2(\mu^2 + \frac{2\mu^2}{|\mathcal{B}_t|} + \alpha \gamma)$.\\
Since $\mu$ satisfies $\mu \leq \frac{\alpha}{6\nu}$
\begin{eqnarray*}
\alpha - 6\mu \nu &\geq& 0 \\
3\alpha - 6\mu \nu &\geq& 2\alpha \\
\frac{3}{\alpha} &\geq& \frac{2}{\alpha - 2\mu \nu}\\
\gamma = \frac{3\mu^2}{\alpha|\mathcal{B}_t|} &\geq& \frac{2\mu^2}{|\mathcal{B}_t|(\alpha - 2\mu \nu)}\\
\end{eqnarray*}

\begin{equation}
  \gamma \geq \frac{2\mu^2}{|\mathcal{B}_t|(\alpha - 2\mu \nu)} \Leftrightarrow \gamma \geq  \frac{\bigg(\frac{2\mu^2}{|\mathcal{B}_t|} + \gamma(1-\alpha)\bigg)}{1 - 2 \mu \nu}
\end{equation}

Because 
\begin{equation}
  \gamma =  \frac{\bigg(\frac{2\mu^2}{|\mathcal{B}|} + \gamma(1-\alpha)\bigg)}{1 - 2 \mu \nu} \Leftrightarrow  \gamma = \frac{2\mu^2}{|\mathcal{B}_t|(\alpha - 2\mu \nu)}
\end{equation}

Also since $\mu$ satisfies $\mu \leq \frac{\nu|\mathcal{B}_t|}{3 \delta^2(|\mathcal{B}_t| + 5)}$

It implies that : 
\begin{eqnarray*}
  \mu (1 + \frac{5}{|\mathcal{B}_t|}) &\leq& \frac{\nu}{3\delta^2} \\
  \mu^2 (1 + \frac{5}{|\mathcal{B}_t|}) &\leq& \frac{\mu\nu}{3\delta^2} \\
  \mu^2 + \frac{2\mu^2}{|\mathcal{B}_t|}  + \frac{3\mu^2}{|\mathcal{B}_t|}&\leq& \frac{\mu\nu}{3\delta^2} \\
  \mu^2 + \frac{2\mu^2}{|\mathcal{B}_t|}  + \alpha\gamma&\leq& \frac{\mu\nu}{3\delta^2} \\
  3\delta^2( \mu^2 + \frac{2\mu^2}{|\mathcal{B}_t|}  + \alpha\gamma )&\leq& \mu\nu \\
  1 - 2\mu \nu + 3\delta^2( \mu^2 + \frac{2\mu^2}{|\mathcal{B}_t|}  + \alpha\gamma )&\leq& 1 - 2\mu \nu +\mu\nu \\
  1 - 2\mu \nu + K &\leq& 1 - \mu \nu
\end{eqnarray*}

So in conclusion : 
\begin{equation}
  1 - 2\mu \nu + K \leq 1 - \mu \nu
\end{equation}

In summary we have : 

\newpage

\vspace*{100cm}
\begin{eqnarray*}
    \mathbb{E}(\|\Tilde{\theta}_{t+1}\|^2) + \gamma G_{t+1} \leq (1-\mu \nu) \cdot \Bigg(\mathbb{E}(\|\Tilde{\theta}_{t}\|^2) + \gamma G_{t}\Bigg) + \frac{4\mu^2}{|\mathcal{B}_t|} C_2
\end{eqnarray*}
Iterating recursion above we get : 
\begin{eqnarray*}
    \mathbb{E}(\|\Tilde{\theta}_{t+1}\|^2) \leq \mathbb{E}(\|\Tilde{\theta}_{t+1}\|^2) + \gamma G_{t+1} &\leq& (1-\mu \nu) \cdot \Bigg(\mathbb{E}(\|\Tilde{\theta}_{t}\|^2) + \gamma G_{t}\Bigg) + \frac{4\mu^2}{|\mathcal{B}_t|} C_2\\
    &\leq& (1-\mu \nu)^{t+1}\cdot \Bigg(\mathbb{E}(\|\Tilde{\theta}_{0}\|^2) + \gamma G_{0}\Bigg) + \frac{4\mu^2}{|\mathcal{B}_t|} C_2 \sum_{k=0}^{t} (1-\mu \nu)^{k}\\
    &\leq& (1-\mu \nu)^{t+1}\cdot \Bigg(\mathbb{E}(\|\Tilde{\theta}_{0}\|^2) + \gamma G_{0}\Bigg) + \frac{1}{1 - (1-\mu \nu)}\frac{4\mu^2}{|\mathcal{B}_t|}  C_2 \\
    &\leq& (1-\mu \nu)^{t+1}\cdot \Bigg(\mathbb{E}(\|\Tilde{\theta}_{0}\|^2) + \gamma G_{0}\Bigg) + \frac{4\mu}{\nu|\mathcal{B}_t|}  C_2 
\end{eqnarray*}

That implies : 
\begin{equation}
 \mathbb{E}(\|\Tilde{\theta}_{t+1}\|^2) \leq (1-\mu \nu)^{t+1}\cdot \Bigg(\mathbb{E}(\|\Tilde{\theta}_{0}\|^2) + \gamma G_{0}\Bigg) + \frac{4\mu}{\nu|\mathcal{B}_t|}  C_2
\end{equation}

So that : 
\begin{equation}
    \limsup_{t \rightarrow +\infty}\mathbb{E}(\|\theta^{*} - \theta_{t+1}\|^2) = O(\frac{\mu}{|\mathcal{B}_t|}  C_2) = O(  \mu\cdot\sigma^2_{in}/|\mathcal{B}_t|)
\end{equation}

%%%%%%%%%%%%%%%%%%%%%%%%%%%%%%%%%%%%%%%%%%%%%%%%%%%%%%%%%%%%%%%%%%%%%%%%%%%%%%%
%%%%%%%%%%%%%%%%%%%%%%%%%%%%%%%%%%%%%%%%%%%%%%%%%%%%%%%%%%%%%%%%%%%%%%%%%%%%%%%

\end{document}